\documentclass[final]{cvpr}
\usepackage{etoolbox}
\usepackage[ruled,vlined]{algorithm2e}
\AtBeginEnvironment{algorithm}{\SetArgSty{textrm}}
\SetKwFor{For}{for (}{) $\lbrace$}{$\rbrace$}
\makeatletter
\patchcmd{\@algocf@start}
  {-1.5em}
  {0pt}
  {}{}
\makeatother

\SetCommentSty{mycommfont}

\usepackage[export]{adjustbox}
\usepackage{booktabs}
\usepackage{multirow}
\usepackage{multicol}
\usepackage{times}
\usepackage{epsfig}
\usepackage{graphicx}
\usepackage{amsmath}
\usepackage{amssymb}
\usepackage{enumitem}
\usepackage[table]{xcolor}
\usepackage{subcaption}
\usepackage{arydshln}

\graphicspath{{images/}}
\newcommand{\was}[1]{}
\def\cvprPaperID{2036} 
\def\confYear{CVPR 2022}

\newcommand{\calL}{{\cal L}}
\newcommand{\calM}{{\cal M}}

\newcommand{\calO}{{\cal O}}

\newcommand{\calS}{{\cal S}}

\newcommand{\bq}{{\bf q}}

\newcommand{\bt}{{\bf t}}

\DeclareMathOperator{\Sim}{sim}

\newcommand{\bbq}{\overline{\bq}}
\newcommand{\bbt}{\overline{\bt}}

\usepackage[pagebackref=true,breaklinks=true,letterpaper=true,colorlinks,bookmarks=false]{hyperref}
\begin{document}

\title{Templates for 3D Object Pose Estimation Revisited:\\
Generalization to New Objects and Robustness to Occlusions}
\author{
	{Van Nguyen Nguyen$^{1}$, \quad Yinlin Hu$^{2}$, \quad Yang Xiao$^{1}$, \quad Mathieu Salzmann$^{2}$, \quad Vincent Lepetit$^{1}$} \\
	{$^{1}$LIGM, Ecole des Ponts, Univ Gustave Eiffel, CNRS, France}\\
	{$^{2}$CVLab, EPFL, Switzerland}\\
	{\tt\small\{van-nguyen.nguyen, yang.xiao, vincent.lepetit\}@enpc.fr }\\
	{\tt\small\{yinlin.hu, mathieu.salzmann\}@epfl.ch}
}

\maketitle


\begin{abstract}
  We present a method that can recognize new objects and estimate their 3D pose in RGB images even under partial occlusions. Our method requires neither a training phase on these objects nor real images depicting them, only their CAD models. It relies on a small set of training objects to learn local object representations, which allow us to locally match the input image to a set of ``templates'', rendered images of the CAD models for the new objects. In contrast with the state-of-the-art methods, the new objects on which our method is applied can be very different from the training objects. As a result, we are the first to show generalization without retraining on the LINEMOD and Occlusion-LINEMOD datasets.  Our analysis of the failure modes of previous template-based approaches further confirms the benefits of local features for template matching. We outperform the state-of-the-art template matching methods on the LINEMOD, Occlusion-LINEMOD and T-LESS datasets. Our source code and data are publicly available at \href{https://github.com/nv-nguyen/template-pose}{https://github.com/nv-nguyen/template-pose}.
\end{abstract}

\vspace*{-3.5mm}
\section{Introduction}
\label{sec:introduction}
3D object pose estimation has significantly improved  over the past decade in terms of both robustness and accuracy~\cite{kehl-iccv17-ssd6d,rad-iccv17-bb8,tekin-cvpr18-realtimeseamlesssingleshot,li-eccv18-deepim,zakharov-iccv19-dpod}. In particular, the robustness to partial occlusions has greatly increased~\cite{peng-cvpr19-pvnet,hu-cvpr19-segmentationdriven6dobjectposeestimation,oberweger-eccv18-makingdeepheatmapsrobust}, and the need for large amounts of real annotated training images has been relaxed thanks to domain transfer~\cite{baek-cvpr20-weaklysuperviseddomainadaptation}, domain randomization~\cite{tremblay-18-deepobjectposeestimation,labbe-eccv20-cosypose,sundermeyer-cvpr20-multipathlearning}, and self-supervised learning~\cite{sundermeyer-ijcv20-augmentedautoencoders} techniques that leverage synthetic images for training.

\begin{figure}[ht]
\newlength{\teaserheight}
\setlength\teaserheight{1.6cm}
\centering
\setlength\lineskip{1.5pt}
\setlength\tabcolsep{1.5pt} 
{\footnotesize
\begin{tabular}{c}
\begin{tabular}{
>{\centering\arraybackslash}m{1.4cm}
>{\centering\arraybackslash}m{\teaserheight}
>{\centering\arraybackslash}m{\teaserheight}
>{\centering\arraybackslash}m{\teaserheight}
>{\centering\arraybackslash}m{\teaserheight}
}
& \multicolumn{2}{c}{Training objects} & \multicolumn{2}{c}{New objects}\\
& w/o occlusion & w/ occlusion & w/o occlusion & w/ occlusion\\ 
Query image &
\includegraphics[height=\teaserheight, ]{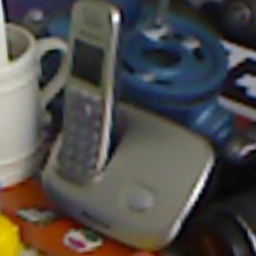}&
\includegraphics[height=\teaserheight, ]{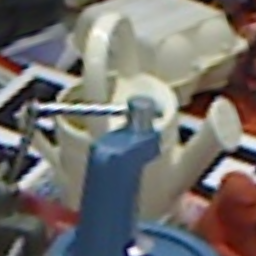}&
\includegraphics[height=\teaserheight, ]{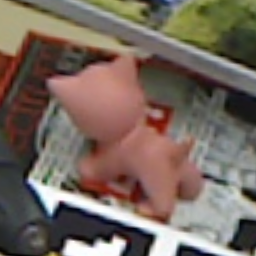}&
\includegraphics[height=\teaserheight, ]{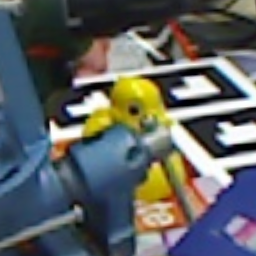}\\
\end{tabular}\\
\begin{tabular}{
>{\centering\arraybackslash}m{1.4cm}
>{\centering\arraybackslash}m{\teaserheight}
>{\centering\arraybackslash}m{\teaserheight}
>{\centering\arraybackslash}m{\teaserheight}
>{\centering\arraybackslash}m{\teaserheight}
}
Template recovered &
\includegraphics[height=\teaserheight, ]{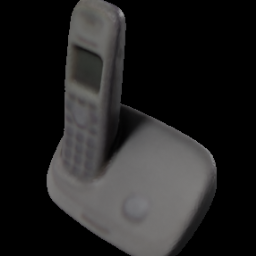}&
\includegraphics[height=\teaserheight, ]{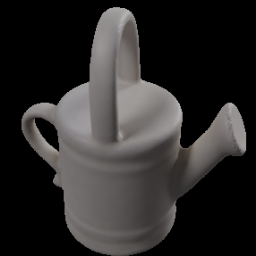}&
\includegraphics[height=\teaserheight, ]{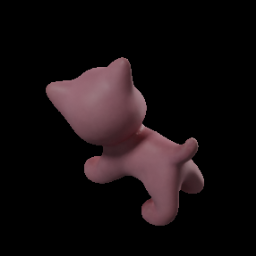}&
\includegraphics[height=\teaserheight, ]{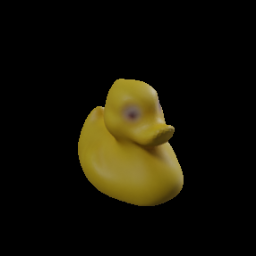}
\end{tabular}
\\
\begin{tabular}{
>{\centering\arraybackslash}m{1.4cm}
>{\centering\arraybackslash}m{\teaserheight}
>{\centering\arraybackslash}m{\teaserheight}
>{\centering\arraybackslash}m{\teaserheight}
>{\centering\arraybackslash}m{\teaserheight}
}
Query image &
\includegraphics[height=\teaserheight, ]{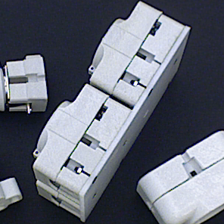}&
\includegraphics[height=\teaserheight, ]{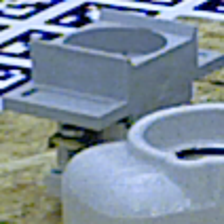}&
\includegraphics[height=\teaserheight, ]{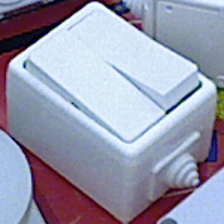}&
\includegraphics[height=\teaserheight, ]{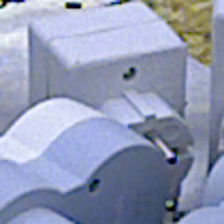}\\
\end{tabular}\\
\begin{tabular}{
>{\centering\arraybackslash}m{1.4cm}
>{\centering\arraybackslash}m{\teaserheight}
>{\centering\arraybackslash}m{\teaserheight}
>{\centering\arraybackslash}m{\teaserheight}
>{\centering\arraybackslash}m{\teaserheight}
}
Template recovered &
\includegraphics[height=\teaserheight, ]{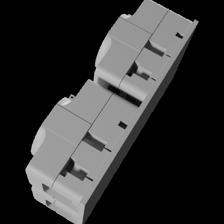}&
\includegraphics[height=\teaserheight, ]{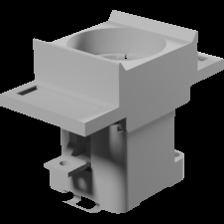}&
\includegraphics[height=\teaserheight, ]{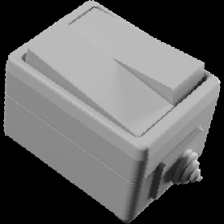}&
\includegraphics[height=\teaserheight, ]{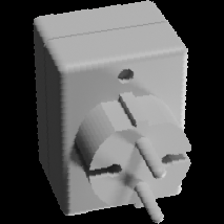}
\end{tabular}
\end{tabular}}
\vspace{-4mm}
\caption{Our method can estimate the 3D pose of new objects in query images by matching them with templates created from their 3D models. \textbf{These new objects can be very different from the  ones, and can be partially occluded in the query images.}}

\vspace{-6mm}
\end{figure}

Nevertheless, the use of image-based 3D object pose estimation remains limited in the industry, despite its huge potential for robotics and augmented reality. Scalable industrial applications would, for example, require the ability to handle arbitrary, previously-unseen objects without retraining and with access only to the objects' CAD models, thus saving both training and data capture time. While a few works have already tackled this challenging task~\cite{sundermeyer-cvpr20-multipathlearning, pitteri-accv20-3dobjectdetection,wohlhart-cvpr15-learningdescriptors,balntas-iccv17-poseguidedrgbdfeaturelearning}, most of them impose some additional constraints by assuming \was{for example} that the new objects belong to a known category~\cite{Wang_2019_NOCS}, remain similar to the training ones as in the T-LESS dataset~\cite{sundermeyer-cvpr20-multipathlearning}, or have prominent corners~\cite{pitteri-accv20-3dobjectdetection}. 

By contrast, template-based approaches~\cite{wohlhart-cvpr15-learningdescriptors,balntas-iccv17-poseguidedrgbdfeaturelearning} offer the promise of generalizing to arbitrary new objects by learning an image embedding used to match the input image to a series of templates generated from their CAD models. Unfortunately, their use with new objects has been demonstrated only anecdotally, and we show in our experiments that these methods struggle in this challenging scenario, particularly in the presence of occlusions. We indeed notice that the global representations used in \cite{wohlhart-cvpr15-learningdescriptors,balntas-iccv17-poseguidedrgbdfeaturelearning} to compare the input image to the CAD-generated templates have two limitations. First, they generalize poorly to new objects in the presence of a cluttered background, and result in inaccurate pose estimation even for uniform background. Furthermore, they are ill-suited to handle occlusions.

These observations motivate us to keep the 2D structure of the images for a template-based approach. More precisely, given a small set of training objects, we learn local features that can be used to reliably match real images and synthetical templates. Relying on local features allows us to discard the background: While the object's mask in the input image is not available at run-time, we can use the template's mask, thus solving the first limitation of global representations. Note that using the template's mask to instead remove the background in the real image before computing the image global representation requires us to recompute the input image representation for each template, which would result in very slow matching.

As will be shown by our experiments, using local features also results in much more accurate poses. This can be explained by the fact that we do not use pooling operations, which remove critical information about the poses, especially for new objects. Finally, yet another advantage is that our method can be robust to partial occlusions. To do so, we introduce a measure to evaluate the similarity between two images that explicitly takes into account the object's mask in the template and the possible occlusions in the query image.

We demonstrate the benefits of our approach on the LINEMOD~\cite{hinterstoisser-accv12-modelbasedtrainingdetection}, Occlusion-LINEMOD~\cite{brachmann-eccv14-learning6dobjectposeestimation}, and T-LESS~\cite{hodan-wacv17-tless} datasets. It consistently outperforms previous works~\cite{wohlhart-cvpr15-learningdescriptors, balntas-iccv17-poseguidedrgbdfeaturelearning, sundermeyer-eccv18-implicit3dorientationlearning, sundermeyer-cvpr20-multipathlearning} on new objects by a large margin. In summary, our contributions are:
\begin{itemize}[noitemsep]
\vspace*{-2mm}
\setlength\itemsep{0em}
    \item A failure-case analysis of previous template-based methods when testing on new objects;
    \item A method that can predict the pose of new objects from their CAD models, without training on these objects nor restricting these objects to be similar to the training ones;
    \item A method robust to occlusions even in the challenging scenario when objects are both new and occluded.
\end{itemize}
\vspace*{-3.5mm}

\section{Related Work}
\label{sec:relatedwork}
Our goal is to develop a method able to estimate the 3D pose of previously-unseen objects while having access only to their 3D model. It should be noted that early approaches to 3D pose estimation already targeted this goal~\cite{lowe-ai87-threedimensionalobjectrecognition}. However, these approaches, based on image edges and object contours, proved to be very fragile. As discussed below, with the use of deep learning, methods have become much more robust but typically require many training images. 

\vspace*{-3.5mm}
\paragraph{Pose estimation for known objects.}
Many 3D object pose estimation methods use a deep model trained on real images or synthetic renderings of these objects, \cite{kehl-iccv17-ssd6d,rad-iccv17-bb8, li-eccv18-deepim,tekin-cvpr18-realtimeseamlesssingleshot,li-iccv19-cdpn,zakharov-iccv19-dpod,park-iccv19-pix2pose,hu-cvpr20-singlestage6dobjectposeestimation}. Some also show remarkable robustness to partial occlusions of the objects~\cite{oberweger-eccv18-makingdeepheatmapsrobust, peng-cvpr19-pvnet, hu-cvpr19-segmentationdriven6dobjectposeestimation}. Such an approach however requires long expensive training and data acquisition/generation time, which we would like to avoid. For example, the state-of-the-art method~\cite{labbe-eccv20-cosypose} on standard benchmarks~\cite{hodan-eccv20-bopchallenge2020on} requires almost a day on 32~GPUs for training. While some works have attempted to reduce the burden of registering real images by learning to generate new images from real ones~\cite{park-eccv20-neuralobjectlearning}, their cost remains too cumbersome for many practical applications.

\vspace*{-2.0mm}
\paragraph{Category-level pose estimation.} 
One way to avoid retraining on new object instances is to consider object categories, and train a model on target categories that will generalize to new instances of these categories~\cite{zhou2018starmap, Wang_2019_NOCS}. While such an approach can \was{indeed} be useful in some applications, such as scene understanding, in many others, the new objects do not belong to a known category. 
By contrast, our approach generalizes to new objects that bear no similarity in shape with the known objects used to train the initial model.

\vspace{-2.0mm}
\paragraph{Unseen object pose estimation.} 
\cite{wohlhart-cvpr15-learningdescriptors} proposed to learn discriminative representations of templates, which are images of objects associated with the corresponding 3D poses. Pose estimation could then be achieved by matching the input image against these templates in an image-retrieval manner. In this context, \cite{balntas-iccv17-poseguidedrgbdfeaturelearning} then showed how to obtain more discriminative representations. While the ability to consider unseen objects by using their 3D models seems to be the motivation for these works, this was only superficially demonstrated, and our experiments show that these methods perform poorly on unseen objects.

More recently, \cite{sundermeyer-cvpr20-multipathlearning} proposed an extension of \cite{sundermeyer-eccv18-implicit3dorientationlearning} to generalize to unseen objects. This method introduces a novel architecture with multiple decoders to adapt to different object types. While their results indeed show  generalization to unseen objects, these objects must remain similar to the training ones. As a consequence, this method has been demonstrated only on the T-LESS dataset, which depicts different kinds of electrical appliances that bear strong visual similarities. 

In any event, as we will discuss in detail in Section~\ref{sec:discuss_global}, these methods rely on a global representation of the templates. We will show that our local representation-based framework has significant advantages in terms of generalization to new objects and of robustness to occlusions. 

\cite{pitteri-accv20-3dobjectdetection} also considers local representations but in a way that is very different from us: \cite{pitteri-accv20-3dobjectdetection} learns to detect specific 2D object locations in the image together with a descriptor for each such location to match them with 3D points on the object's 3D model. This matching, however, is done independently for each location, making it highly ambiguous, and resulting in a combinatorial matching cost and frequent failures. By contrast, we extract local representations in a grid structure and learn to match all local input and template representations jointly. To achieve this, we rely on contrastive learning, which we discuss below.

A different and interesting take was proposed in \cite{xiao-bmvc19-posefromshape}, where the embedding of the object's 3D model was used as input in addition to the input image to predict the 3D pose. However, this work considers only pose regression and assumes the object is already known in order to use the right 3D model.

\vspace{-2.5mm}
\paragraph{Contrastive learning.} 
Given a collection of images, contrastive learning aims to learn an embedding space where similar images are close to each other while dissimilar ones are far apart. \cite{hjelm2018learning,Wu2018UnsupervisedFL,Oord2018infoNCE,tian2019contrastive,He2020moco,chen2020simCLR} leverage unlabeled images and strong data augmentation to learn powerful image features that achieve results competitive with those of supervised learning on various downstream tasks.



In our context, \cite{Xiao2020PoseContrast} exploits a form of contrastive learning, leveraging the pose labels to learn a pose-aware embedding space for class-agnostic 3D object pose estimation.  One limitation of \cite{Xiao2020PoseContrast} is that different objects can be mixed with each other in the embedding space, thus making it impossible to recognize the correct object instance from the input image. Moreover, like \cite{xiao-bmvc19-posefromshape}, \cite{Xiao2020PoseContrast} does not attempt to recognize the object.

By contrast, \cite{wohlhart-cvpr15-learningdescriptors, balntas-iccv17-poseguidedrgbdfeaturelearning} rely on contrastive learning to learn an embedding space that is variant to both the object pose and the object instance. To this end, they rely on a triplet loss for learning object-discriminative features, together with a pairwise loss for pose-discriminative features. 
Similarly, we use contrastive learning to extract a discriminative feature representation, but we show that the InfoNCE~\cite{Oord2018infoNCE} loss is the most simple and effective choice. Our experiments also show that most of the performance of our method in terms of generalization and robustness to occlusions come from our use of local representations.

\vspace*{-2.mm}
\section{Method}
\label{sec:method}

Our goal is to recognize new objects in color images and predict their 3D poses. We do this by matching the color image of the object with a set of templates. A template is a rendered image of a 3D model in some 3D pose. For each new object, the template set contains many templates, rendered from different views sampled around its 3D model. As the templates are annotated with the object's identity and pose, the method returns the identity and pose of the template most similar to the input image.

The challenge then is to measure the similarity between templates and input images. This should be done reliably despite that no real images of the new objects have been seen beforehand, the objects can be partially occluded, the lighting differs between the templates and the real images, and the object's background is cluttered in the real images.

In this work, motivated by the better repeatability and robustness to occlusions of local representations compared to global ones, we measure the similarity between an input image and a template based on local image features extracted using a deep model. We train this model using pairs made of a real image and a synthetic image from a small set of training objects. Note that these training objects can be very different in appearance from the new objects. 

We start this section with an analysis of the limits of global representations in Section~\ref{sec:discuss_global}. We then detail in Section~\ref{sec:training} our training procedure. It relies on a similarity measure that compares the local features of real images and synthetic templates. At run-time, we use an extended version of this similarity function that explicitly estimates which local features in the input image are occluded and discards them. We discuss this in Section~\ref{sec:occ}.  Finally, we detail how we generate the templates in Section~\ref{sec:rendering}.


\vspace{-1mm}
\subsection{Motivation and Analysis}
\label{sec:discuss_global}
\begin{figure}[!t]
\newlength{\plotheight}
\setlength\plotheight{2.25cm}
\centering
\setlength\lineskip{2.5pt}
\setlength\tabcolsep{2.5pt} 
{\small
\begin{tabular}{
c
>{\centering\arraybackslash}m{\plotheight}
>{\centering\arraybackslash}m{\plotheight}
>{\centering\arraybackslash}m{\plotheight}
}
&Seen objects & Unseen objects & After masking\\
\rotatebox{90}{\cite{balntas-iccv17-poseguidedrgbdfeaturelearning}}&
\frame{\includegraphics[height=\plotheight, ]{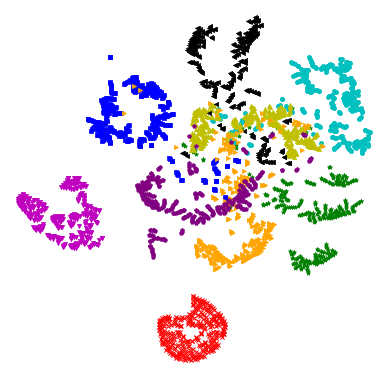}}&
\frame{\includegraphics[height=\plotheight, ]{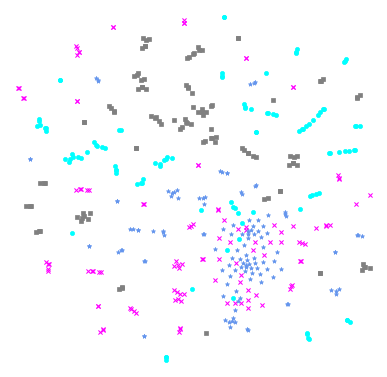}}&
\frame{\includegraphics[height=\plotheight, ]{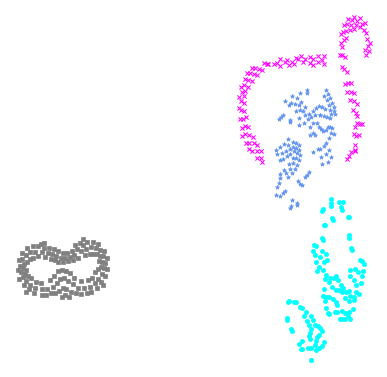}}\\[0em]
\rotatebox{90}{$\!\!\!\!\!\!\!\!\!\!\!\!$Our method}&
\frame{\includegraphics[height=\plotheight, ]{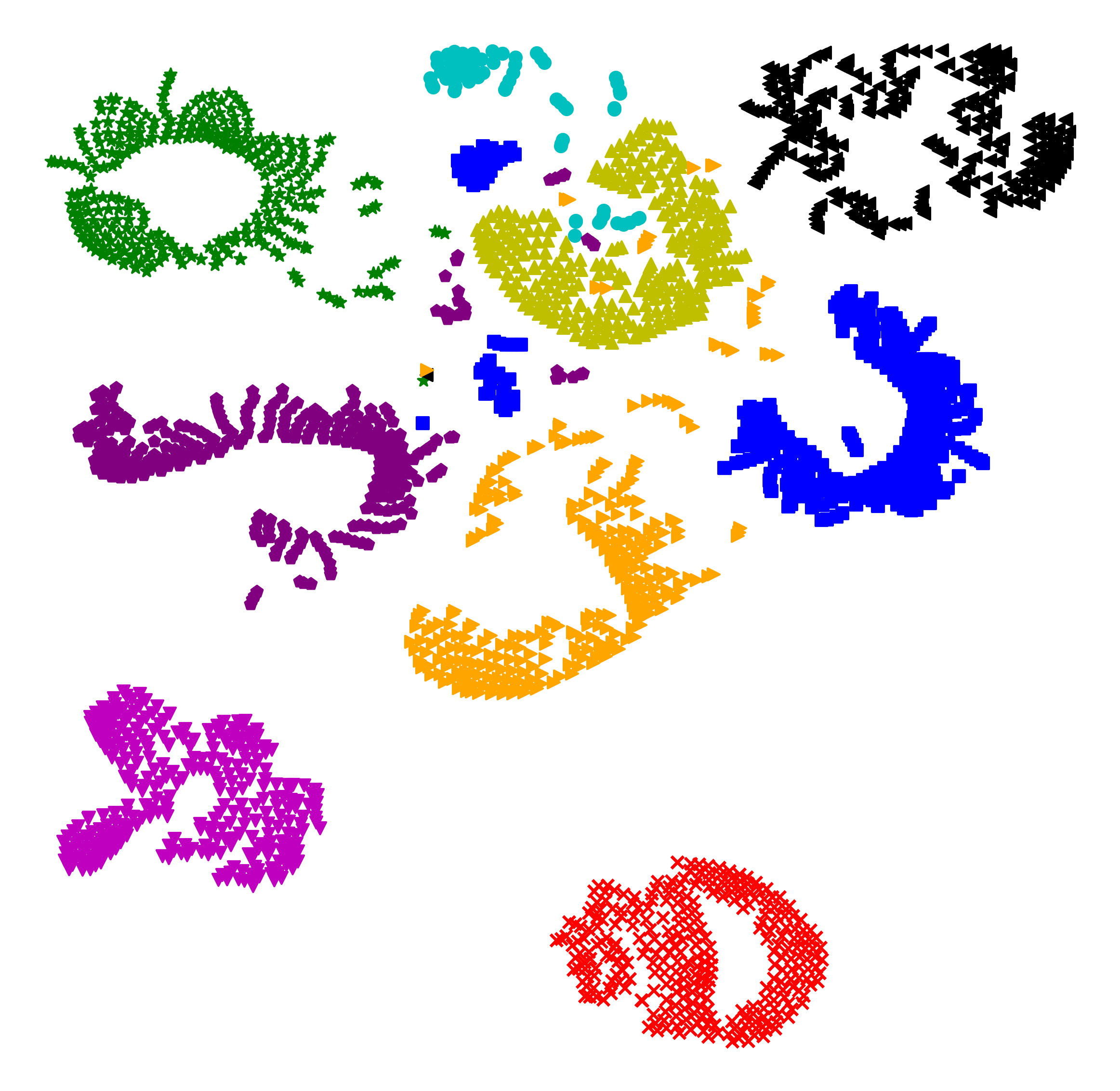}}&
\frame{\includegraphics[height=\plotheight, ]{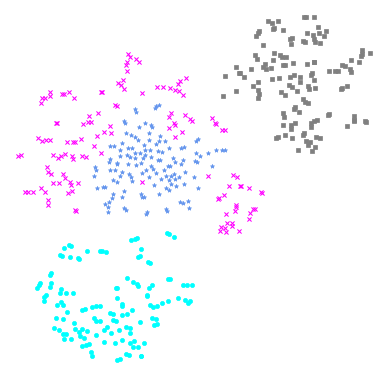}}&
\frame{\includegraphics[height=\plotheight, ]{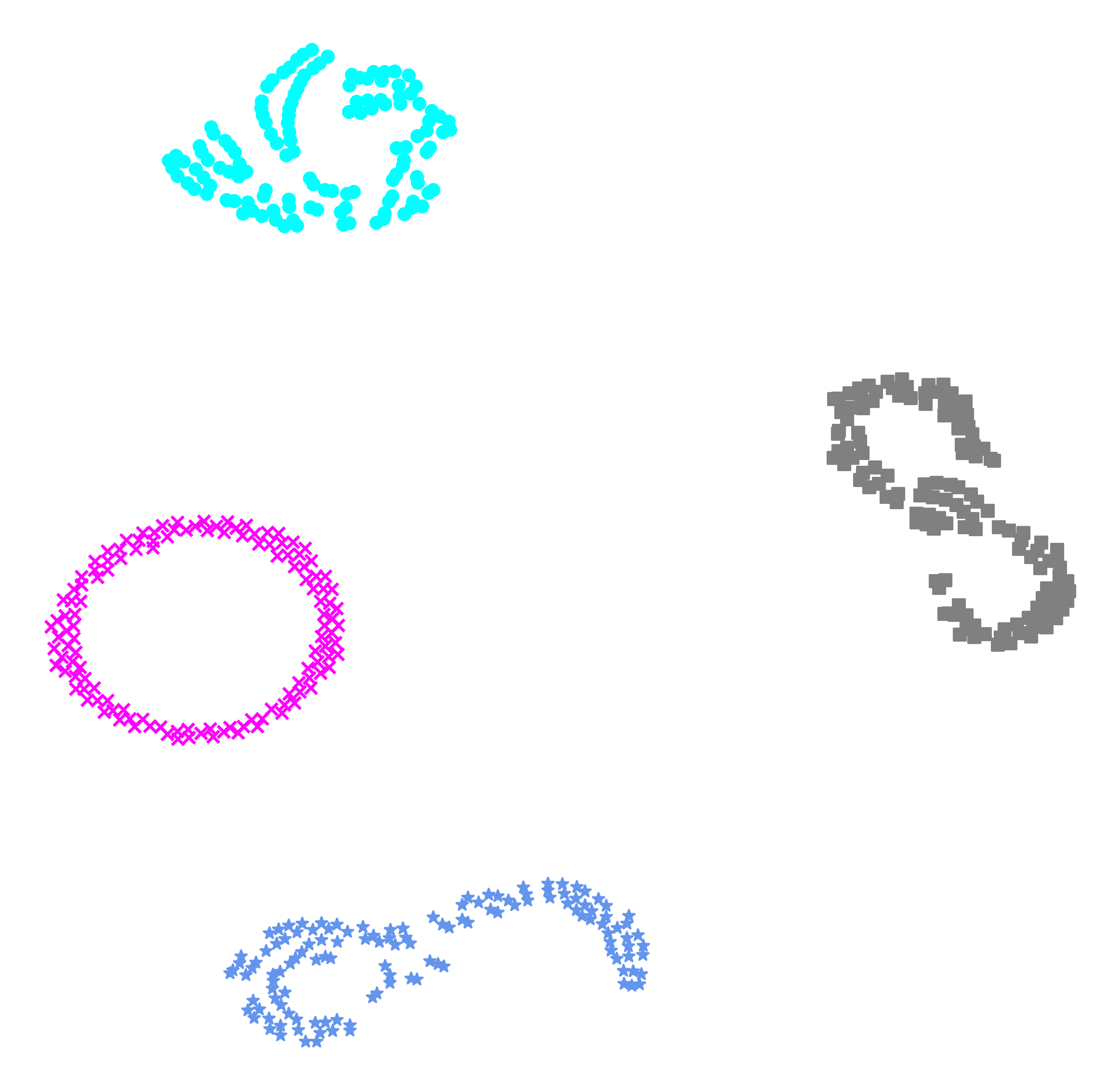}}\\
\end{tabular}
\begin{tabular}{c@{\;\;\;}c}
\includegraphics[height=0.3\plotheight, ]{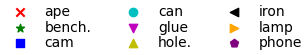}&
\includegraphics[height=0.3\plotheight, ]{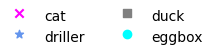}\\
\end{tabular}
}
\vspace{-4mm}
\caption{\textbf{Understanding the influence of background on different image representations,} with T-SNE visualizations of the image representations learned by \cite{balntas-iccv17-poseguidedrgbdfeaturelearning}~(first row) and by our method~(second row) for real images of LINEMOD objects. For a given column, all the plots have the same scale for comparison. 
}
\label{fig:tsne}
\end{figure}
Here, we present two experiments that point out the main drawbacks of global representations in template matching when working with unseen objects.
\vspace{-3.5mm}
\begin{figure*}[!h]
    \begin{center}
    \includegraphics[width=1\linewidth]{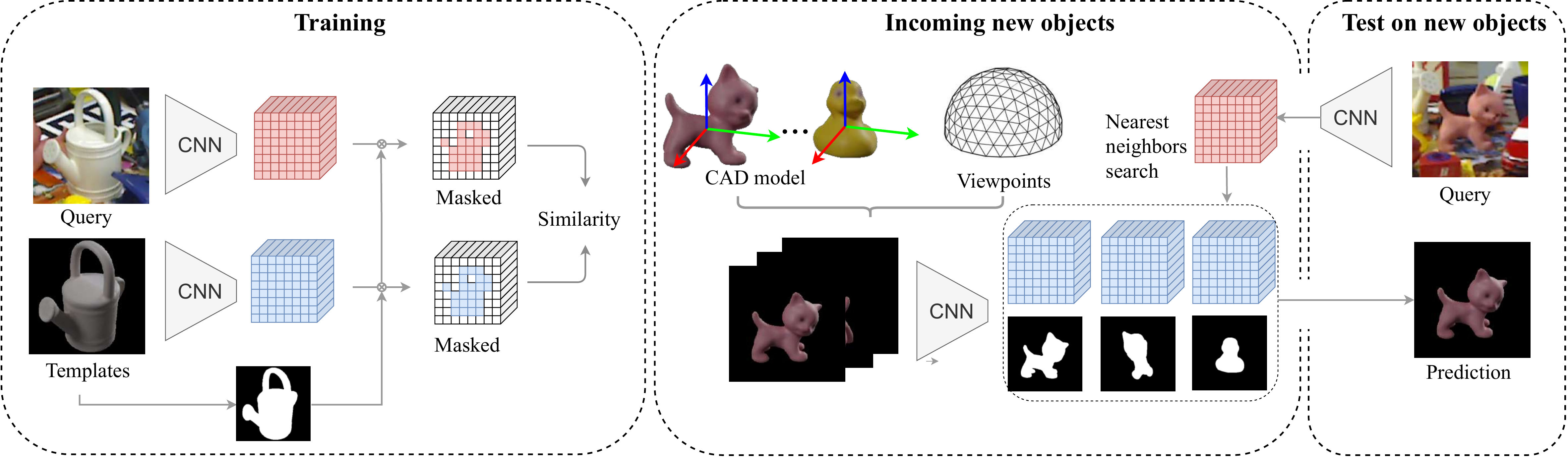}
    \end{center}
    \vspace{-6mm}
    \caption{
        \label{fig:framework}
        {\bf At training time}, we use pairs made of a real image and a synthetic template to train a network to compute local features, from which the similarity between the two images can be predicted. {\bf At run-time}, we apply this network to images of objects not seen during training to compute their local features. We can then retrieve the object pose by matching the image against the database of templates.}
\end{figure*}

\subsubsection{Cluttered Background}
\label{sec:bg}
A first drawback of global representations is their poor ability to represent unseen objects on cluttered backgrounds. To show this, we plot in Figure~\ref{fig:tsne} the t-SNE visualization \cite{van2008visualizing} of the global representations learned by \cite{balntas-iccv17-poseguidedrgbdfeaturelearning} and local representations learned by our method for real images of training and new objects of the LINEMOD dataset.

The first column of Figure~\ref{fig:tsne} shows that both representations manage to cluster the images of each training object together, despite the fact that the images of the objects are captured with a cluttered background. The second column shows that global representations of \cite{balntas-iccv17-poseguidedrgbdfeaturelearning} cannot  disentangle the images of unseen objects, while our representations can. To better understand the reason behind this, we remove the background in the images by replacing it with a uniform color using the ground-truth object masks. As shown in the third column, the representations are now disentangled. This shows the influence of the background on the global representations for unseen objects, and that our representations are robust to cluttered backgrounds.

\vspace{-4mm}
\subsubsection{Pose Discrimination}
\label{sec:pose_disc}
\vspace{-1mm}
A second drawback of global representations is their poor reliability when matching the real image of an unseen object with the synthetic template for the corresponding 3D pose, even when the object identity is known and the background is uniform. This can be explained by the fact that the pooling layers remove important information. This information loss appears to be compensated by the rest of the architecture for the training objects, but this compensation does not generalize to unseen objects.


To show this, we visualize in the supplementary material the correlation between pose distances and representation distances for unseen objects, as done in \cite{wohlhart-cvpr15-learningdescriptors, balntas-iccv17-poseguidedrgbdfeaturelearning}. While both representations result in a strong correlation for training objects, this correlation is lost when considering unseen objects for the global representations but not for ours. Even without background, the correlation is still very low for global representations~\cite{balntas-iccv17-poseguidedrgbdfeaturelearning}.
\subsection{Framework}
\label{sec:training}

In each training iteration, we sample $N$ positive pairs, where pair $i$ is composed of a real image $\bq_i$ depicting a training object and of a synthetic template $\bt_i$ of the same object in a similar 3D pose. Following~\cite{wohlhart-cvpr15-learningdescriptors}, we deem the two viewpoints  similar if the angle between them is less than 5 degrees. All the pairs composed by a real image and a synthetic image of different objects or dissimilar poses~(larger than 5 degrees) are defined as negative pairs.

\vspace{-4mm}
\paragraph{Triplet loss.} \cite{wohlhart-cvpr15-learningdescriptors} proposed a metric learning approach based on the intuition that the distance between feature descriptors for positive pairs should be closer in the learnt embedding space than negative pairs. To learn this property, \cite{wohlhart-cvpr15-learningdescriptors} used a training loss $\calL = \calL_{triplet} + \calL_{pair}$ where: 
\begin{itemize}[noitemsep]
\vspace*{-2mm}
\setlength\itemsep{0em}
    \item $\calL_{triplet}$ is the triplet term, which allows the network to learn features such that the distance in the learned embedding space between the positive pairs $\Delta_{+}^{(i)}$ is lower than the distance between the negative pairs  $\Delta_{-}^{(i)}$ within the limits of the margin $m$. This triplet term is defined as
\begin{equation}\label{loss:triplet}
\calL_{triplet}=\sum_{i=1}^N \max \left(0, 1-\frac{\Delta_{+}^{(i)}}{\Delta_{-}^{(i)}+m}\right)
\end{equation}
    \item $\calL_{pair}=\sum_{i=1}^N \Delta_{+}^{(i)}$ is the pairwise term, to minimise distances between two images of identical poses but different viewing conditions.
\end{itemize}
\vspace*{-2mm}
\cite{balntas-iccv17-poseguidedrgbdfeaturelearning} made an extension of this work by proposing a triplet loss which focuses only on learning object-discriminative features while using a pairwise loss to learn an embedding space analogous to the pose differences.

While these two losses work well, we experimentally show that the recent standard contrast loss InfoNCE~\cite{Oord2018infoNCE} is the most simple and effective choice.

 
\vspace{-3mm}
\paragraph{InfoNCE loss.}
For each real image $\bq_i$, we also create $N-1$ negative pairs by combining it with synthetic templates $\bt_k$ of other pairs in the current batch, with $1\leq k\leq N, k\neq i$. Altogether, this yields $N$ positive pairs and $(N-1)\times N$ negative pairs for each batch. We train our model to maximize the agreement between the representations of samples in positive pairs, while minimizing that of negative pairs with the InfoNCE loss function~\cite{Oord2018infoNCE}:
\begin{equation}\label{loss:infoNCE}
    \calL = - \sum_{i=1}^N \log \frac{\exp{(\Sim(\bbq_i, \bbt_i)/\tau)}}{\sum_{k=1}^N 1_{[k\neq i]}\exp{(\Sim(\bbq_i, \bbt_k)/\tau)}} \> ,
\end{equation}
where $\Sim(\bbq, \bbt)$ measures the similarity between the local image features $\bbq$ and $\bbt$ computed by the deep model for real image $\bq$ and template $\bt$, and $\tau=0.1$, is a temperature parameter. As shown in Figure~\ref{fig:framework}, $\bbq$ and $\bbt$ retain a grid structure
and are 3-tensors. In practice, their dimensions depend on the size of the input image, ranging from $25\times25\times C$ to $28\times 28 \times C$, with $C=16$.

\vspace{-3mm}
\paragraph{Local feature similarity.}
While previous works on contrastive learning~\cite{Oord2018infoNCE,tian2019contrastive,Misra2020PIRL,caron2020swav,He2020moco,chen2020mocov2,chen2020simCLR,chen2020simCLRv2} focused mostly on image classification and define the similarity metric $\Sim(.,.)$ using a global representation of the two images, we found such a representation to only classify well either known objects or images with a clean background, as discussed in Section~\ref{sec:bg}. To effectively handle new objects and complex backgrounds, we use a metric based on a pairwise comparison of the local features in $\bbq$ and $\bbt$. Specifically, we define
\begin{equation}
    \Sim(\bbq, \bbt) = \frac{1}{|\calM|} \sum_l \calM^{(l)} \calS\left(\bbq^{(l)}, \bbt^{(l)}\right) \> ,
\label{eq:sim}
\end{equation}
where  $\calS$ is a local similarity metric, $\calM$ is a 2D binary visibility mask for template $\bt$, and index $l$ indicates a 2D grid location. $\bbq^{(l)}$ and $\bbt^{(l)}$ are thus local features of dimension $C$. Considering the template mask allows us to discard the background in the real image. Note that the mask does not account for possible occlusions in the real image as it corresponds to the object's silhouette in the template. Occlusions will be considered in the next subsection. As a local similarity metric $\calS$, we use the cosine similarity
\begin{equation}
\calS\left(\bbq^{(l)}, \bbt^{(l)}\right) = \frac{\bbq^{(l)}}{||\bbq^{(l)}||} \cdot \frac{\bbt^{(l)}}{||\bbt^{(l)}||} \> ,
\end{equation}
We empirically observed that measuring the similarity as the opposite of the L1 and L2 norms of the differences yields the same performance as the cosine similarity. 

\subsection{Run-time and Robustness to Occlusions}
\label{sec:occ}
\begin{figure}[!t]
\newlength{\imageheight}
\setlength\imageheight{1.43cm}
\centering
\setlength\lineskip{1.5pt}
\setlength\tabcolsep{1.5pt} 
{\small
\begin{tabular}{c}
\begin{tabular}{
>{\centering\arraybackslash}m{1.0cm}
>{\centering\arraybackslash}m{0.5cm}
>{\centering\arraybackslash}m{\imageheight}
>{\centering\arraybackslash}m{\imageheight}
>{\centering\arraybackslash}m{\imageheight}
>{\centering\arraybackslash}m{\imageheight}
}
& &  & Template 1 & Template 2 & Template 3\\ 
& & &
\includegraphics[height=\imageheight, ]{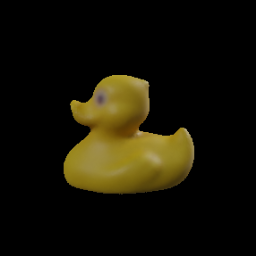}&
\includegraphics[height=\imageheight, ]{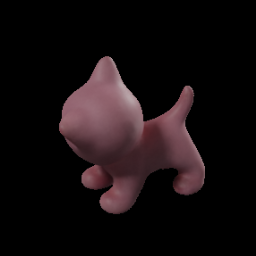}&
\includegraphics[height=\imageheight, ]{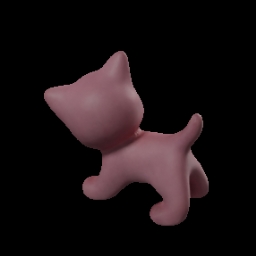}\\
\end{tabular}\\
\begin{tabular}{
>{\centering\arraybackslash}m{1.0cm}
>{\centering\arraybackslash}m{0.5cm}
>{\centering\arraybackslash}m{\imageheight}
>{\centering\arraybackslash}m{\imageheight}
>{\centering\arraybackslash}m{\imageheight}
>{\centering\arraybackslash}m{\imageheight}
}
\raisebox{0.0\height}{\rotatebox[origin=c]{0}{ \parbox{1.2cm}{w/o $\calO$}}} & \raisebox{0.0\height}{\rotatebox[origin=c]{90}{ \parbox{1.cm}{Query}}} &
\includegraphics[height=\imageheight, ]{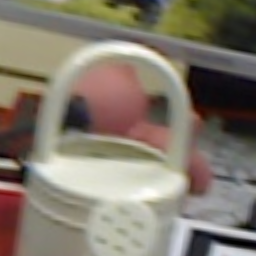}&
\includegraphics[height=\imageheight, ]{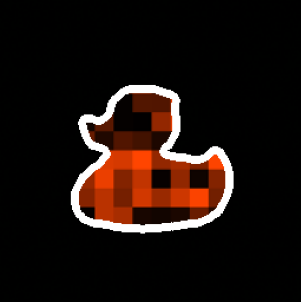}&
\includegraphics[height=\imageheight, ]{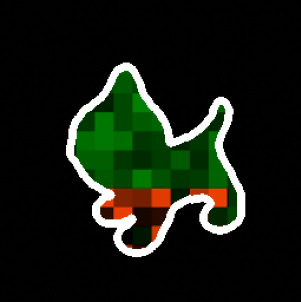}&
\includegraphics[height=\imageheight, ]{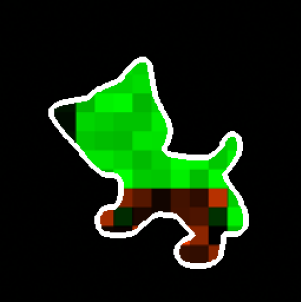}\\
\end{tabular}\\
\begin{tabular}{
>{\centering\arraybackslash}m{1.0cm}
>{\centering\arraybackslash}m{0.5cm}
>{\centering\arraybackslash}m{\imageheight}
>{\centering\arraybackslash}m{\imageheight}
>{\centering\arraybackslash}m{\imageheight}
>{\centering\arraybackslash}m{\imageheight}
}
\raisebox{0.0\height}{\rotatebox[origin=c]{0}{\parbox{1.2cm}{w/ $\calO$}}} & \raisebox{0.0\height}{\rotatebox[origin=c]{90}{ \parbox{1.0cm}{Query}}} &
\includegraphics[height=\imageheight, ]{images/methods/occlusion_aware/occlusion_aware_query.png}&
\includegraphics[height=\imageheight, ]{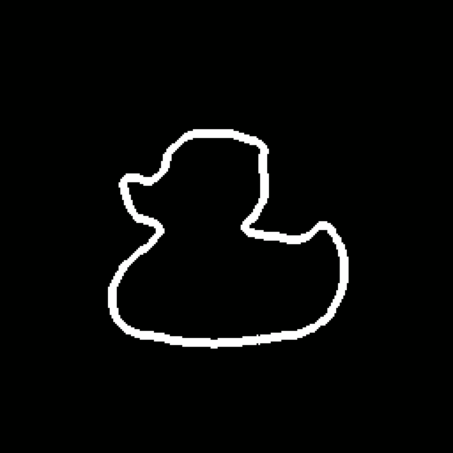}&
\includegraphics[height=\imageheight, ]{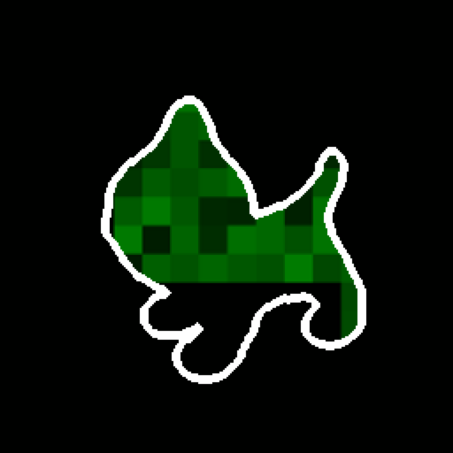}&
\includegraphics[height=\imageheight, ]{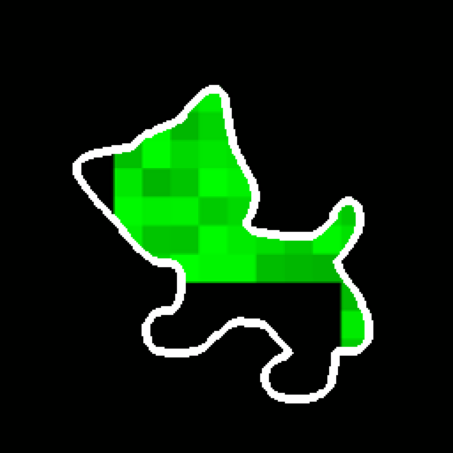}\\
\end{tabular}\\
\begin{tabular}{@{\;\;\;\;}
>{\arraybackslash}m{5cm}
}
\includegraphics[height=0.48\imageheight, ]{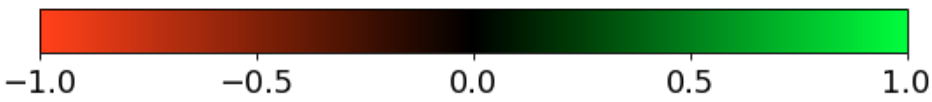}\\
\end{tabular}\\
\end{tabular}
}

\vspace{-4mm}
\caption{{\bf Illustration of feature similarity} when not using the occlusion mask  $\calO$~(second row) and when using it. As discussed in Section~\ref{sec:occ}, using $\calO$ allows  “turning off”  the possible  occluded  local  features in the similarity score. }
\label{fig:occlusion}
\end{figure}

At run-time, given a real query image $\bq$, we retrieve the most similar template in a template set. To be robust to occlusions that can occur in the query image, we modify $\Sim(\bbq, \bbt)$ as:
\begin{equation}
    \Sim^*(\bbq, \bbt) = \frac{1}{|\calM|} \sum_l \calM^{(l)} \calO^{(l)} \calS\left(\bbq^{(l)}, \bbt^{(l)}\right) \> ,
\label{eq:sim_runtime}
\end{equation}
where $\calO^{(l)} = 1_{\calS(\bbq^{(l)}, \bbt^{(l)})>\delta}$  with $\delta$ a threshold applied to the cosine similarity to ``turn off'' the occluded local features as shown in Figure \ref{fig:occlusion}. In practice, we set this threshold $\delta=0.2$ through ablation study. Note that Eq.~\eqref{eq:sim_runtime} can be written as the element-wise product $\odot$ and can be computed efficiently with:
\begin{equation}
\Sim^*(\bbq, \bbt) = \frac{1}{|\calM|} (\calM\odot\calO\odot\calS) \> .
\label{eq:sim_runtime2}
\end{equation}

\vspace*{-3.5mm}
\subsection{Template Creation}
\label{sec:rendering}
On LINEMOD~\cite{hinterstoisser-accv12-modelbasedtrainingdetection} and  Occlusion-LINEMOD~\cite{brachmann-eccv14-learning6dobjectposeestimation} datasets, we follow the protocol of \cite{wohlhart-cvpr15-learningdescriptors} to sample the synthetic templates.  More precisely, the viewpoints are defined by starting with a regular icosahedron and recursively subdividing each triangle into 4 smaller triangles. After applying this subdivision two times and removing the lower half-sphere, we end up with 301 templates per object. 


On T-LESS~\cite{hodan-wacv17-tless}, we follow the protocol of \cite{sundermeyer-cvpr20-multipathlearning} by using a dense regular icosahedron with 2'536 viewpoints and 36 in-plane rotations for each rendered image. Altogether, this yields 92'232 templates per object. Besides, we also show our results with a coarser regular icosahedron with 602 viewpoints, which results 21'672 templates per object. 

We use BlenderProc~\cite{denninger2019blenderproc} to generate templates with realistic rendering for both settings. 
\vspace*{-2.5mm}

\section{Experiments}
\label{sec:experiments}
In this section, we first describe the experimental setup~(Section~\ref{sec:experimental setup}). Then, we compare quantitatively and qualitatively our method with previous works~\cite{wohlhart-cvpr15-learningdescriptors, balntas-iccv17-poseguidedrgbdfeaturelearning, sundermeyer-eccv18-implicit3dorientationlearning, sundermeyer-cvpr20-multipathlearning} on both training (or seen) and unseen objects of the LINEMOD~(LM)~\cite{hinterstoisser-accv12-modelbasedtrainingdetection}, Occlusion-LINEMOD~(O-LM)~\cite{brachmann-eccv14-learning6dobjectposeestimation} and T-LESS~\cite{hodan-wacv17-tless} datasets (Section~\ref{sec:main_results}). Finally, we provide an ablation study for investigating the effectiveness of our method with different parameters and failure cases of our method~(Sections~\ref{sec:ablation} and \ref{sec:failures}).


\subsection{Experimental Setup}
\label{sec:experimental setup}
\paragraph{Data processing.} For the LM and O-LM datasets, as there are no standard splits to evaluate the robustness of RGB-based methods on unseen objects, we propose three different splits created from the order of the object ids. The new, or unseen objects for each of these splits are:
\begin{itemize}[noitemsep]
\setlength\itemsep{0em}
    \item Split~\#1: \textbf{Ape}, Benchvise, Camera, \textbf{Can};
    \item Split~\#2: \textbf{Cat}, \textbf{Driller}, \textbf{Duck}, \textbf{Eggbox};
    \item Split~\#3: \textbf{Glue}, \textbf{Holepuncher}, Iron, Lamp, Phone.
\end{itemize}
The other objects from LM are used for training the model. The objects with names in \textbf{bold} in the lists above often are occluded in O-LM. Note that O-LM  is only used for testing, as we do not need to see occlusions during the training time. Moreover, to understand the performance gap between objects that are seen or unseen during the training, we also evaluate the methods on seen objects. To do so, we keep 10\% of the real images of training objects under unseen poses for testing purposes. Table~\ref{tab:datasets} details the different splits.

On T-LESS~\cite{hodan-wacv17-tless}, we follow the evaluation protocol of \cite{sundermeyer-cvpr20-multipathlearning} by training only on objects 1-18 under randomized backgrounds of SUN397 \cite{xiao2010sun} and testing on the complete T-LESS primesense test set. More
details about training set of T-LESS can be found in the supplementary material.

\begin{table}[!t]
    \addtolength{\tabcolsep}{-2pt}
    \centering
    \scalebox{0.73}{
    \begin{tabular}{@{}c c c c c c@{}}
    \hline
    Split & Training & Seen LM & Seen O-LM  & Unseen LM  & Unseen O-LM\\
    \hline
    \#1 & 9'954 & 981 & 6'832 & 4'848 & 2'377\\
    \#2 & 9'928 & 981 & 4'490 & 4'874 & 4'719\\
    \#3 & 8'850 & 872 & 7'096 & 6'061 & 2'113\\
    \hline
    \end{tabular}}
    \vspace*{-3mm}
    \caption{{\bf Dataset splits for LM and O-LM.} For each split, we provide the numbers of real images in the training set and in four test sets. }
    \vspace*{-2mm}
    \label{tab:datasets}
\end{table}

\vspace{-3mm}
\paragraph{Evaluation metrics.} 
For the LM and O-LM datasets, the pose error is measured by the angle between the two positions on the viewing half-sphere. We also treat the “Eggbox" and “Glue" objects as symmetric around the z-axis as done in \cite{wohlhart-cvpr15-learningdescriptors, balntas-iccv17-poseguidedrgbdfeaturelearning}.

In the case of known object pose estimation, the recognition score is almost 100\% on LM and O-LM. Previous works \cite{wohlhart-cvpr15-learningdescriptors, balntas-iccv17-poseguidedrgbdfeaturelearning} that focused on known objects thus only evaluate the pose error without considering whether the retrieved object is actually correct. In the case of unseen objects, we found that retrieving correctly both pose and class is important as the model can still get correct poses but from another object. Therefore, we propose using the Acc15 metric, which measures how often the pose error is less than 15 degrees \emph{and} the predicted object class is correct. We also report the pose error in the supplementary material. 

As most objects in T-LESS~\cite{hodan-wacv17-tless} are symmetric, we report the recall under the Visible Surface Discrepancy~($\text{err}_{vsd}$) metric at $\text{err}_{vsd}<0.3$ with tolerance $\tau=20 mm$ and $>10\%$ object visibility as done in \cite{sundermeyer-eccv18-implicit3dorientationlearning, sundermeyer-cvpr20-multipathlearning}. Unless otherwise stated in previous works~\cite{sundermeyer-eccv18-implicit3dorientationlearning, sundermeyer-cvpr20-multipathlearning}, only templates of the same object are used at testing time (in other words, the class of the object is assumed to be known before testing). Please note that for the evaluation on the T-LESS dataset, we also predict the translation by using the same formula “projective distance estimation" of SSD-6D \cite{kehl-iccv17-ssd6d} as done in~\cite{sundermeyer-eccv18-implicit3dorientationlearning, sundermeyer-cvpr20-multipathlearning}. This translation is deduced from the retrieved template and the input bounding box of query image. More details can be found in the supplementary material.

\vspace{-3mm}
\paragraph{Implementation details. } 
\label{ImplementationDetails}

For a fair comparison, in the evaluation on LM and O-LM, we consider two different backbones: (i) ``Base'' -- the simple backbone used in \cite{wohlhart-cvpr15-learningdescriptors,balntas-iccv17-poseguidedrgbdfeaturelearning}; (ii) ResNet50 -- the standard backbone used in recent contrastive learning methods~\cite{He2020moco}. We reimplemented \cite{wohlhart-cvpr15-learningdescriptors,balntas-iccv17-poseguidedrgbdfeaturelearning} to get quantitative results in both seen and unseen objects. Our implementations  get very similar performance when evaluated on the same data as the original papers on seen objects (see Table~\ref{tab:linemod_full}), validating our reimplementation.

We also follow \cite{wohlhart-cvpr15-learningdescriptors,balntas-iccv17-poseguidedrgbdfeaturelearning} when testing with the ``Base'' backbone by using the same input image of size 64$\times$64. While testing with ResNet50, we use a larger input size of 224$\times$224.  In both settings, we slightly change the architecture by removing all the pooling, FC layers and then replace them by two $1\times1$ convolution layers to output the desired local feature of size 16. As done in~\cite{wohlhart-cvpr15-learningdescriptors, balntas-iccv17-poseguidedrgbdfeaturelearning}, we use the ground-truth pose to crop the input image at the center of objects and do not consider in-plane rotation (more details can be found in the supplementary material). On the T-LESS dataset, we use the same backbone ResNet50 and crop the input image with ground-truth bounding box as done in \cite{sundermeyer-eccv18-implicit3dorientationlearning, sundermeyer-cvpr20-multipathlearning}.

For both evaluations, we train our networks using Adam with an initial learning rate of 1e-2 for the ``Base'' backbone and of 1e-4 for ResNet50. Training takes less than 5h for all splits on a single V100 GPU when training on LM~\cite{hinterstoisser-accv12-modelbasedtrainingdetection} and around 12h when training on T-LESS~\cite{hodan-wacv17-tless}.

\subsection{Comparison with the State of the Art}
\label{sec:main_results}
\subsubsection{LINEMOD and Occluded-LINEMOD Results}
\begin{table*}[!t]
\addtolength{\tabcolsep}{-1pt}
\centering
    \scalebox{.76}{
    \begin{tabular}{l l l l| 
    c  c  c | c |
    c  c  c | c |
    c  c  c | c |
    c  c  c | c }
	\toprule
	\multirow{2}{*}{\bf Method} &
	\multirow{2}{*}{\bf Backbone} &
	\multirow{2}{*}{\bf Features} &
	\multirow{2}{*}{\bf Loss} &
	\multicolumn{4}{c|}{\textbf{Seen LM}}&
	\multicolumn{4}{c|}{\textbf{Seen O-LM}}&
	\multicolumn{4}{c|}{\textbf{Unseen LM}}& 
	\multicolumn{4}{c}{\textbf{Unseen O-LM}}\\
	
    \cmidrule(lr){5-8} \cmidrule(lr){9-12}\cmidrule(lr){13-16}\cmidrule(lr){17-20}
	
	& & & & ~\#1 & ~\#2 & ~\#3 & Avg. & ~\#1 & ~\#2 & ~\#3 & Avg. & ~\#1 & ~\#2 & ~\#3 & Avg. & ~\#1 & ~\#2 & ~\#3 & Avg. \\
	\midrule
	
	\rowcolor{gray!10}
	\cite{wohlhart-cvpr15-learningdescriptors} & Base~\cite{wohlhart-cvpr15-learningdescriptors} & Global & \cite{wohlhart-cvpr15-learningdescriptors} & 87.0 & 83.1 & 85.1 & 85.0 & 19.2 & 23.1 & 15.0 & 19.1 & 13.2 & 15.5 & 18.2 & 15.2 &  9.3 & 5.1 & 5.1 & 6.5\\
	
	\cite{wohlhart-cvpr15-learningdescriptors} & Base~\cite{wohlhart-cvpr15-learningdescriptors} & Global & Eq.~\eqref{loss:infoNCE} & 95.2 & 95.3 & 95.4 & 95.3 & 19.6 & 25.3 & 16.1 & 20.3 & 13.3 & 17.0 & 20.5 & 16.9 & 8.2 & 6.4 & 6.7 & 7.1\\
	
	\rowcolor{gray!10}
	\cite{balntas-iccv17-poseguidedrgbdfeaturelearning} & Base~\cite{wohlhart-cvpr15-learningdescriptors}& Global & \cite{balntas-iccv17-poseguidedrgbdfeaturelearning} & 89.2 & 85.4 & 83.3 & 86.3 & 18.3 & 21.9 & 17.6 & 19.5 & 14.1 & 16.3 & 19.7 & 16.7 &  8.2 & 7.5 & 7.6 & 7.8\\
	
	\cite{balntas-iccv17-poseguidedrgbdfeaturelearning} & Base~\cite{wohlhart-cvpr15-learningdescriptors}& Global & Eq.~\eqref{loss:infoNCE} & 96.3 & 95.2 & 96.5 & 96.0 & 18.3 & 23.1 & 15.8 & 19.1 & 11.5 & 17.7 & 17.2 & 15.5 &  7.1 & 6.5 & 6.5 & 6.7\\
	
	\rowcolor{gray!10}
	Ours & Base~\cite{wohlhart-cvpr15-learningdescriptors}& Local & \cite{wohlhart-cvpr15-learningdescriptors} & 84.8 & 85.5 & 86.3 & 85.5 & 50.1 & 51.3 & 42.2 & 47.9 & 69.6 & 63.2 & 46.2 & 59.7 &  35.3 & 34.3 &  44.2 & 37.9\\
    
    Ours & Base~\cite{wohlhart-cvpr15-learningdescriptors}& Local & Eq.~\eqref{loss:infoNCE} & 95.6 & 96.9 & 92.0 & 94.8  & 68.9 &  71.0  & 57.7  & 65.8  &  78.8  & 82.5 & 64.1  & 75.1  &  42.2 & 57.1  & 59.8  & 53.0 \\
    
    \cmidrule(lr){1-20}
    
    \rowcolor{gray!10}
	\cite{wohlhart-cvpr15-learningdescriptors} & ResNet50 ~\cite{he-cvpr16-deepresiduallearning} & Global & Eq.~\eqref{loss:infoNCE} & 98.8 & 96.9 &  98.8 & 98.1 & 66.7 & 73.2 & 62.7 & 67.5 & 42.2 & 43.7 & 49.4 & 45.1 & 22.3 & 22.5 & 45.9 & 29.9\\
	
	\cite{balntas-iccv17-poseguidedrgbdfeaturelearning} & ResNet50 ~\cite{he-cvpr16-deepresiduallearning}& Global & Eq.~\eqref{loss:infoNCE}  & 96.9 & 97.1 & 94.5 & 96.1 & 63.6 & 71.8 & 58.9 & 64.7 & 39.9 & 44.9 & 48.3 & 44.3 & 15.5 & 21.8 & 50.2 & 29.1\\
    
    \rowcolor{gray!10}
    
    Ours & ResNet50 ~\cite{he-cvpr16-deepresiduallearning}& Local & Eq.~\eqref{loss:infoNCE} & \bf 99.3 & \bf 99.0 &  \bf 99.2 & \bf 99.1  & \bf 77.3 & \bf 84.1 & \bf 76.8 & \bf  79.4 & \bf 94.4 & \bf 97.4 & \bf 88.7 &  \bf 93.5  & \bf 71.4 & \bf 72.7 & \bf 85.3 & \bf  76.3\\
\bottomrule
\end{tabular}} 
    
\vspace*{-2mm}
\caption{\textbf{Comparison of our method with \cite{ wohlhart-cvpr15-learningdescriptors} and \cite{balntas-iccv17-poseguidedrgbdfeaturelearning}} on seen and unseen objects of LM and O-LM under the three different splits detailed at the beginning of Section~\ref{sec:experimental setup}. 
We report $\bf \text{Acc15} \uparrow$, the accuracy of predicting correctly the object identity \emph{and} its pose with an error less than 15 degrees. We are on par on the ``easy'' case and outperform them by a large margin on the 3 other configurations.
Using the InfoNCE loss rather than the loss from \cite{balntas-iccv17-poseguidedrgbdfeaturelearning} brings some improvement, but the main improvement comes from our approach based on local features.}
\label{tab:linemod_full}
\end{table*}

\begin{figure*}

\setlength\imageheight{1.4cm}
\centering
\setlength\lineskip{1.pt}
\setlength\tabcolsep{1.pt} 
{\small
\begin{tabular}{c}
\begin{tabular}{
>{\centering\arraybackslash}m{1.cm}
>{\centering\arraybackslash}m{\imageheight}
>{\centering\arraybackslash}m{\imageheight}
>{\centering\arraybackslash}m{\imageheight}
>{\centering\arraybackslash}m{\imageheight}
>{\centering\arraybackslash}m{\imageheight}
>{\centering\arraybackslash}m{0.2cm}
>{\centering\arraybackslash}m{1.cm}
>{\centering\arraybackslash}m{\imageheight}
>{\centering\arraybackslash}m{\imageheight}
>{\centering\arraybackslash}m{\imageheight}
>{\centering\arraybackslash}m{\imageheight}
>{\centering\arraybackslash}m{\imageheight}
}
 & 
\includegraphics[height=\imageheight, ]{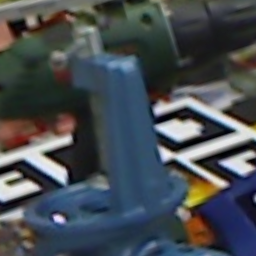} &
\includegraphics[height=\imageheight, ]{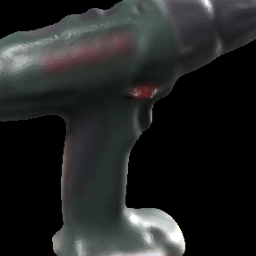}&
\includegraphics[height=\imageheight, ]{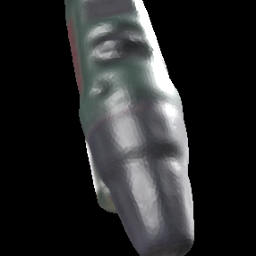}&
\includegraphics[height=\imageheight, ]{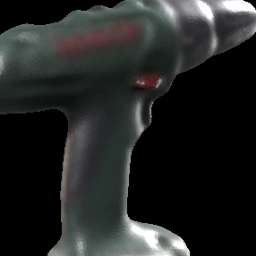}&
\includegraphics[height=\imageheight]{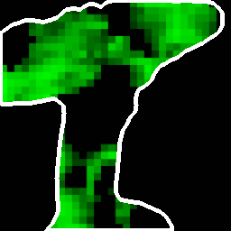}&
&
 &
 \includegraphics[height=\imageheight, ]{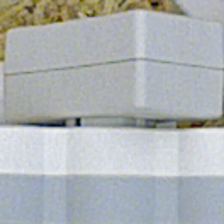} &
\includegraphics[height=\imageheight, ]{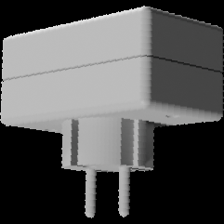}&
\includegraphics[height=\imageheight, ]{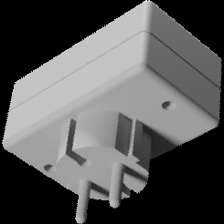}&
\includegraphics[height=\imageheight, ]{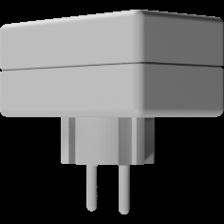}&
\includegraphics[height=\imageheight]{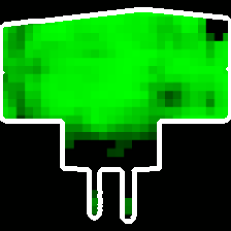}\\

\end{tabular}\\

\begin{tabular}{
>{\centering\arraybackslash}m{1.cm}
>{\centering\arraybackslash}m{\imageheight}
>{\centering\arraybackslash}m{\imageheight}
>{\centering\arraybackslash}m{\imageheight}
>{\centering\arraybackslash}m{\imageheight}
>{\centering\arraybackslash}m{\imageheight}
>{\centering\arraybackslash}m{0.2cm}
>{\centering\arraybackslash}m{1.cm}
>{\centering\arraybackslash}m{\imageheight}
>{\centering\arraybackslash}m{\imageheight}
>{\centering\arraybackslash}m{\imageheight}
>{\centering\arraybackslash}m{\imageheight}
>{\centering\arraybackslash}m{\imageheight}
}
\raisebox{0.0\height}{\rotatebox[origin=c]{90}{\bf \parbox{1.2cm}{Occlusion LINEMOD}}} & 
\includegraphics[height=\imageheight, ]{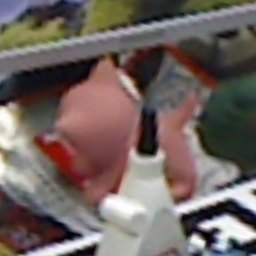} &
\includegraphics[height=\imageheight, ]{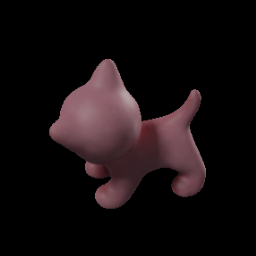}&
\includegraphics[height=\imageheight, ]{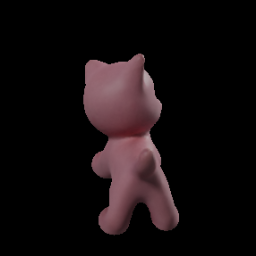}&
\includegraphics[height=\imageheight, ]{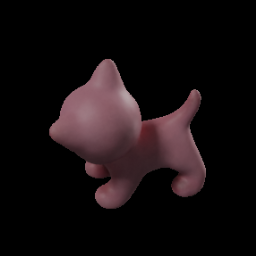}&
\includegraphics[height=\imageheight]{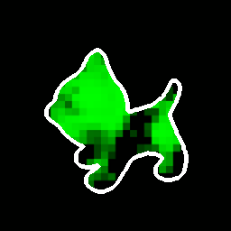}&
&
\raisebox{0.3\height}{\rotatebox[origin=c]{90}{\bf \parbox{1.2cm}{T-LESS}}} 

 &
 \includegraphics[height=\imageheight, ]{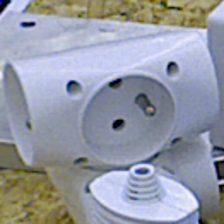} &
\includegraphics[height=\imageheight, ]{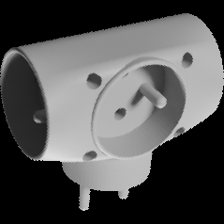}&
\includegraphics[height=\imageheight, ]{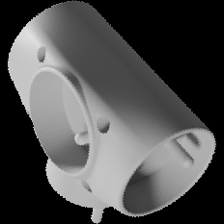}&
\includegraphics[height=\imageheight, ]{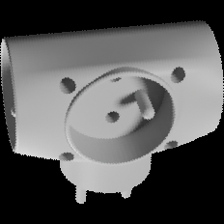}&
\includegraphics[height=\imageheight]{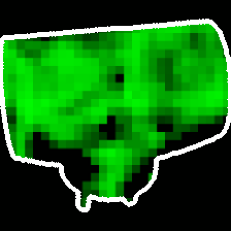}\\
\end{tabular}\\

\begin{tabular}{
>{\centering\arraybackslash}m{1.cm}
>{\centering\arraybackslash}m{\imageheight}
>{\centering\arraybackslash}m{\imageheight}
>{\centering\arraybackslash}m{\imageheight}
>{\centering\arraybackslash}m{\imageheight}
>{\centering\arraybackslash}m{\imageheight}
>{\centering\arraybackslash}m{0.2cm}
>{\centering\arraybackslash}m{1.cm}
>{\centering\arraybackslash}m{\imageheight}
>{\centering\arraybackslash}m{\imageheight}
>{\centering\arraybackslash}m{\imageheight}
>{\centering\arraybackslash}m{\imageheight}
>{\centering\arraybackslash}m{\imageheight}
}
 & 
\includegraphics[height=\imageheight, ]{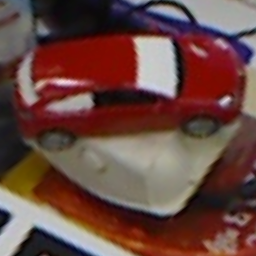} &
\includegraphics[height=\imageheight, ]{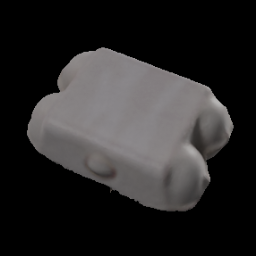}&
\includegraphics[height=\imageheight, ]{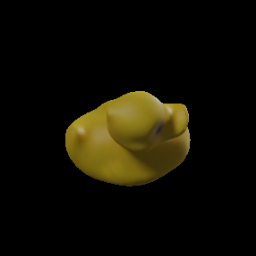}&
\includegraphics[height=\imageheight, ]{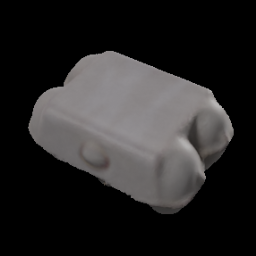}&
\includegraphics[height=\imageheight]{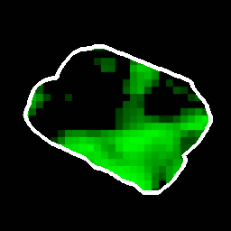}&
&
&
\includegraphics[height=\imageheight, ]{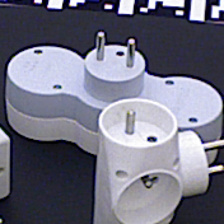} &
\includegraphics[height=\imageheight, ]{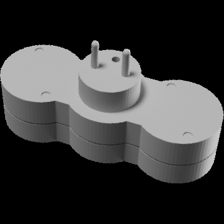}&
\includegraphics[height=\imageheight, ]{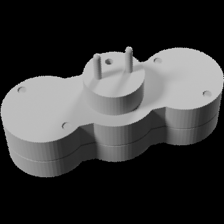}&
\includegraphics[height=\imageheight, ]{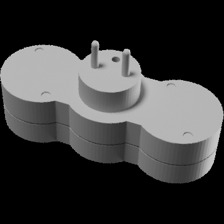}&
\includegraphics[height=\imageheight]{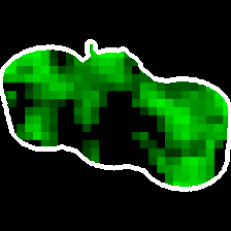}\\
& Query & GT & \cite{balntas-iccv17-poseguidedrgbdfeaturelearning} & Ours & Similarity  &  & & Query & GT & \cite{sundermeyer-cvpr20-multipathlearning} & Ours & Similarity
\end{tabular}
\end{tabular}
}
\vspace{-3mm}
\caption{{\bf Qualitative results on unseen objects} of Occlusion-LINEMOD (left) and T-LESS (right). Our method retrieves the correct template and pose while \cite{balntas-iccv17-poseguidedrgbdfeaturelearning, sundermeyer-cvpr20-multipathlearning} fails on unseen objects, particularly in the presence of occlusion.
}
\end{figure*}
\begin{table}[t]
	\centering
    \resizebox{0.9\linewidth}{!}{
	\begin{tabular}{@{}l c c c c c@{}}
	\toprule
    \multirow{2}{*}{\bf Method} & \multirow{2}{*}{\bf \parbox{1.5cm}{Number templates}} &
    \multicolumn{3}{c}{\textbf{Recall VSD}}\\
	\cmidrule(lr){3-5}                                
	& &\textbf{Obj. 1-18} & \textbf{Obj. 19-30} & \textbf{Avg} \\
	\midrule
	\rowcolor{gray!10} Implicit \cite{sundermeyer-eccv18-implicit3dorientationlearning} & 92K & 35.60 & 42.45 & 38.34\\
	MPL \cite{sundermeyer-cvpr20-multipathlearning} & 92K & 35.25 & 33.17 & 34.42\\
	
	\rowcolor{gray!10} Ours & 92K & \bf 59.62 & \bf 57.75 & \bf 58.87\\
	\midrule
	Ours & 21K & 59.14 & 56.91 & 58.25\\
	
	\bottomrule
	\end{tabular}}
	\vspace{-3mm}	\caption{{\bf Comparison with \cite{sundermeyer-eccv18-implicit3dorientationlearning, sundermeyer-cvpr20-multipathlearning}} on seen objects (obj. 1-18) and unseen objects (obj. 19-31) of T-LESS using the protocol from \cite{sundermeyer-cvpr20-multipathlearning}. Our method significantly outperforms \cite{sundermeyer-eccv18-implicit3dorientationlearning, sundermeyer-cvpr20-multipathlearning} in the same setting.}
    \label{tab:tless}
\end{table}
\vspace{-2mm}
Table~\ref{tab:linemod_full} presents the results of our method compared with previous work~\cite{wohlhart-cvpr15-learningdescriptors,balntas-iccv17-poseguidedrgbdfeaturelearning}. With either the ``Base'' or ResNet50 backbones, our method based on local feature similarities achieves the best overall performance in almost all settings compared to previous methods that compute the feature similarity between global image representations. 
While \cite{wohlhart-cvpr15-learningdescriptors,balntas-iccv17-poseguidedrgbdfeaturelearning} explored carefully designed pairwise and triplet losses for learning an embedding space that is both object-discriminative and pose-discriminative, we find that using the InfoNCE loss as defined in Eq.~\eqref{loss:infoNCE} boosts the performance of all methods, in particular for our method based on local feature similarities.

When the objects are occluded, the  accuracy of \cite{wohlhart-cvpr15-learningdescriptors,balntas-iccv17-poseguidedrgbdfeaturelearning} drops to below 70\% for training objects, while our method can still maintain a relatively high accuracy. This shows the robustness of local image features rather than global image representations that are much more strongly affected by the occlusions. Furthermore, the prediction accuracy of our method on unseen objects is clearly higher than that of previous methods, regardless of the objects being occluded or not. This indicates that matching based on local features is not only robust to occlusions, but also generalizes better to unseen objects. More importantly, this improvement on unseen objects holds still in the presence of occlusions.

\subsubsection{T-Less Results}
\label{tless_results}

In Table~\ref{tab:tless}, we shown that our proposed approach outperforms the state-of-the-art methods~\cite{sundermeyer-eccv18-implicit3dorientationlearning, sundermeyer-cvpr20-multipathlearning} on the T-LESS dataset by a large margin on both seen and unseen objects. While ~\cite{sundermeyer-cvpr20-multipathlearning} carefully designed single-encoder-multi-decoder network that allows sharing a latent space for all objects and having each decoder only reconstructs views of a single object, we find that using our method and Info-NCE loss is much more simple but also boosts significantly the performance in the same setting.


\begin{table}
	\centering
    \scalebox{.70}{
	\begin{tabular}{@{}l |c c |c c c c |c c| c@{}}
	\toprule
    & \bf Ape& \bf Can& \bf Cat& \bf Driller&\bf Duck&\bf Egg$^*$& \bf Glue$^*$& \bf Hole.& \bf Avg\\
	\midrule                                  
	\rowcolor{gray!10} \cite{wohlhart-cvpr15-learningdescriptors} & 16.6 & 28.0 & 1.5 & 8.2 & 11.5 & 68.8 & 67.7 & 22.1 & 29.9\\
	\cite{balntas-iccv17-poseguidedrgbdfeaturelearning} & 12.6 & 18.4 & 9.0 & 16.7 & 7.8 & 53.7 & 60.3 & 40.1 & 29.1\\
	\rowcolor{gray!10}
	Ours & \bf 53.8 & \bf 89.7 & \bf 45.1 & \bf 84.4 & \bf 87.2 & \bf 76.9 & \bf 89.9 & \bf 83.3 & \bf 76.3 \\
	{} w/o $\calM$ & 13.3 & 1.0 & 10.0 & 1.0 & 80.1 & 7.0 & 80.0 & 1.0 & 24.1\\
	\bottomrule
	\end{tabular}}
	\vspace{-2mm}
	\caption{{\bf Effectiveness of $\calM$}. Comparison of \cite{wohlhart-cvpr15-learningdescriptors, balntas-iccv17-poseguidedrgbdfeaturelearning} and our method with and without using the template mask $\calM$ in the computation of the similarity. Using $\calM$ allows discarding the cluttered background and brings significant improvement on occluded unseen objects. }
    \label{tab:occlusion_linemod}
\end{table} 

\begin{table}[!h]
	\centering
	\rowcolors{2}{gray!10}{white}
    \scalebox{.70}{
	\begin{tabular}{@{}l l c c c c c c c c c@{}}
	\toprule
    Threshold $\delta$ & -0.3 & -0.2 & -0.1 & 0 & 0.1  & 0.2 & 0.3 & w/o $\cal O$\\
	\midrule                                  
	\bf Ape & 54.1 & 53.7 & 54.6 & \bf 54.7 & 54.0 & 53.8 & 53.6 & 53.3\\
	\bf Can & 82.2 & 89.2 & 89.1 & 89.7 & 89.4 & 89.7 & \bf 89.8 & 84.9\\
	\midrule     
	\bf Cat & 46.7 & \bf 47.5 & 46.1 & 45.5 & 46.1 & 45.1 &  46.5 & 45.1\\
	\bf Driller &  83.6 & \bf 84.5 & \bf 84.5 & 83.8 & 84.4 & 84.4 & \bf \bf 84.5 & 81.5\\
	\bf Duck	& 87.1 & 87.1 & \bf 87.8 & 86.7 & 87.3 &   87.2 & 87.0 & 87.3\\
	\bf Egg$^*$ & 76.3 & 75.2 & 74.1 & 75.3 & 75.1 & \bf 76.9 & 76.2 & 72.6\\
	\midrule     
	\bf Glue$^*$& 89.3 & 83.5 & 83.9 &  90.1 & 89.5 & 89.9 & 89.6 & \bf 90.2\\
	\bf Holep. & 83.9 & \bf 85.9 & 83.6 & 82.9 & 83.4 & 83.3 & 82.5 & 81.8\\
	\midrule
	\bf Avg & 75.4 & 75.8 & 75.4 & 76.0 & 76.1 & \bf 76.3 & 76.2  & 74.5\\
	\bottomrule
	\end{tabular}}
	\vspace{-2mm}
	\caption{\textbf{Influence of threshold $\delta$ of Eq.~\eqref{eq:sim_runtime}. } Predicting occlusion mask $\calO$ with  threshold $\delta=0.2$ results on the best performance, particularly on large objects.
	}
    \label{tab:threshold_occlusion_lm}
\end{table} 
\subsection{Ablation Study}
\label{sec:ablation}

We present several ablation evaluations on LINEMOD and Occlusion-LINEMOD.
\vspace{-3mm}
\paragraph{Effectiveness of feature masking.}
Table~\ref{tab:occlusion_linemod} shows the effectiveness of using the template masks $\calM$ in Eq.~\eqref{eq:sim_runtime2} for unseen objects. Removing $\calM$ results in a dramatic degradation for our method on all the three splits. 

\paragraph{Influence of the threshold $\delta$.} 
Table~\ref{tab:threshold_occlusion_lm} shows the influence of the threshold $\delta$ in Eq.~\eqref{eq:sim_runtime} for estimating the occlusion mask $\calO$. Using $\calO$ brings improvements on large objects~(“Can", “Driller", and “Eggbox"). This can be explained by the fact that the occlusions can be very large in O-LM, especially on small objects, as shown in Figure~\ref{fig:cat_heavy}.

\begin{figure}
  \centering
  \begin{tabular}{cc}
  \includegraphics[width=0.45\linewidth]{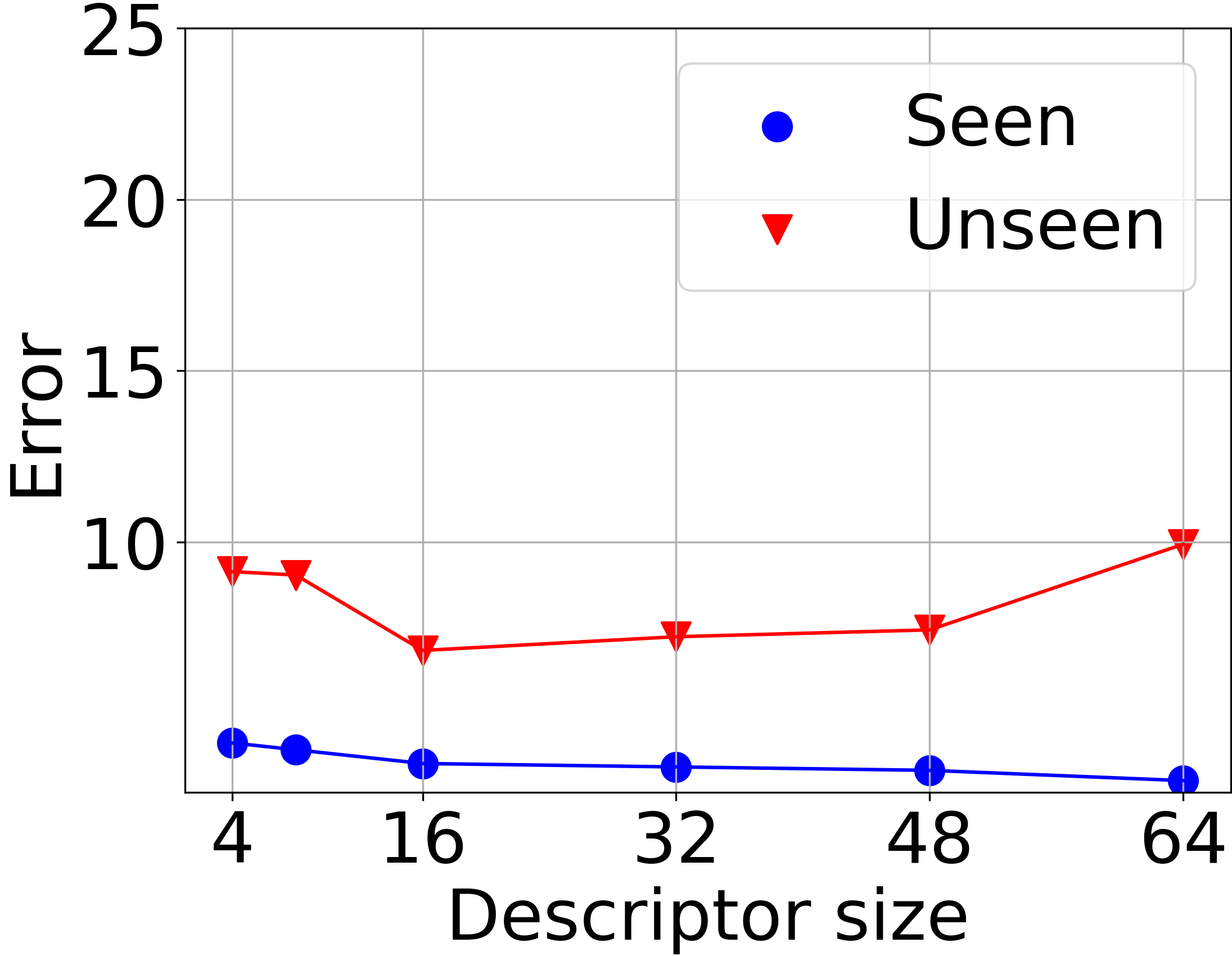} &
    \includegraphics[width=0.45\linewidth]{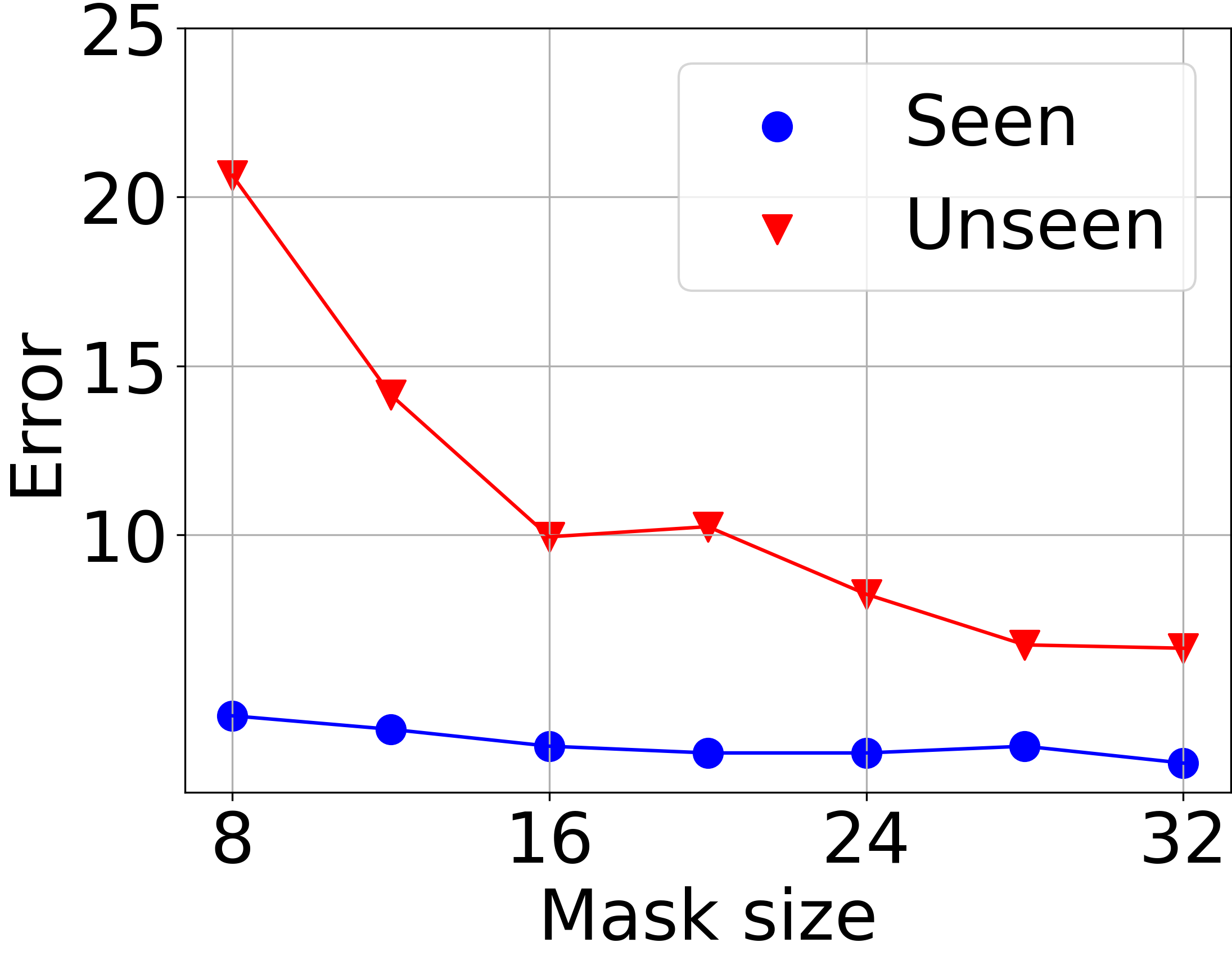}\\
  \end{tabular}
  \vspace{-3mm}
  \caption{\label{fig:mask_size}\textbf{Influence of the local feature dimension $C$ and of the resolution of the local features and masks.} Using a good resolution is much more important than using high-dimensional local features as this allows discarding background more precisely when computing the similarity score.}
\end{figure}

\vspace{-3mm}
\paragraph{Influence of the local feature dimensions.}
Figure~\ref{fig:mask_size} shows the pose error as a function of the dimension $C$ of the local features and of the resolution of the feature maps and masks $\calM$. While $C$ is not a critical value, the resolution is more important, as higher resolution allows discarding the background more precisely. Furthermore, this hyperparameter has a much stronger influence on the performance on the unseen objects compared to the seen objects.

\begin{table}[t]
	\centering
	\resizebox{0.9\linewidth}{!}{
	\begin{tabular}{@{}l r r r c c@{}}
	\toprule
    \multirow{2}{*}{\bf Dataset} & \multirow{2}{*}{\bf \parbox{1.5cm}{Number templates}}
    &\multirow{2}{*}{\bf \parbox{1.3cm}{Features creation}}
    &\multirow{2}{*}{\bf \parbox{1.3cm}{Memory}}
     & 
    \multicolumn{2}{c}{\textbf{Run-time}}\\
	\cmidrule(lr){5-6}                                
	& & & & \textbf{CPU} & \textbf{GPU} \\
	\midrule
	LINEMOD & 1.204 & 0.5 min & 28 MB & 0.15 s & 7.8$\times 10^{-3}$ s\\
	T-LESS & 21.672 & 6 min & 544 MB & 0.84 s & 8.2$\times 10^{-3}$ s\\
    \bottomrule
	\end{tabular}}
	\vspace*{-2.5mm}
	\caption{{\bf Average run-time}  of our method on a single GPU V100 and CPU Intel Xeon.}
    \label{tab:runtime}
\end{table}
\paragraph{Run-time.}
Table~\ref{tab:runtime} provides run-times on CPU and GPU.

\subsection{Failure Cases}
\label{sec:failures}
\begin{figure}[!t]
  \centering \addtolength{\tabcolsep}{-2pt}
  \begin{tabular}{ccc}
  \includegraphics[width=0.3\linewidth]{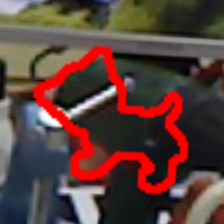} & 
  \includegraphics[width=0.3\linewidth]{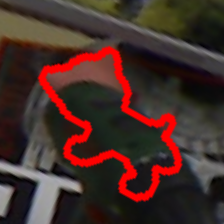} & 
  \includegraphics[width=0.3\linewidth]{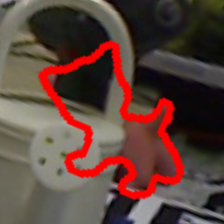}\\
  \end{tabular}
  \vspace{-3mm}
  \caption{The “Cat" object is often barely visible in the test images of Occluded-LINEMOD, resulting in large errors.  }
 \label{fig:cat_heavy}
\end{figure}

When evaluated on O-LM, both our method and \cite{wohlhart-cvpr15-learningdescriptors,balntas-iccv17-poseguidedrgbdfeaturelearning} fail on the “Cat" object. 
As shown in Figure~\ref{fig:cat_heavy}, this object is small and  particularly heavily occluded in this dataset. 

\section{Conclusion}
\label{sec:conclusion}
We have presented an efficient approach to 3D object recognition and pose estimation that can generalize to new objects without the need for retraining and that is robust to occlusions.  Our analysis has shown that a global representation, which discards the grid structure of images, is not robust to clutter and results in inaccurate pose predictions. Our method, based on local representations, has much better properties and can be made robust to occlusions. We hope that our analysis and our new approach will guide the development of more practical systems.

\vspace{0.5cm}
{\small \noindent\textbf{Acknowledgments.} We thank Micha\"el Ramamonjisoa, Tom Monnier, Elliot Vincent and Romain Loiseau for valuable feedback. This research was produced within the framework of Energy4Climate Interdisciplinary Center (E4C) of IP Paris and Ecole des Ponts ParisTech. This research was supported by 3rd Programme d’Investissements d’Avenir [ANR-18-EUR-0006-02]. This action benefited from the support of the Chair “Challenging Technology for Responsible Energy" led by l’X – Ecole polytechnique and the Fondation de l’Ecole polytechnique, sponsored by TOTAL. This work has received funding from the CHISTERA IPALM project and was performed using HPC resources from GENCI–IDRIS 2021-AD011012294R1.}

{\small
\bibliographystyle{ieee_fullname}
\bibliography{bibliography}
}
\section*{}
\newpage
\textbf{\Large{Supplementary Material}}

\section{Pose Discrimination}
\label{sec:pose_discrimination}

\begin{figure}
\setlength\plotheight{2.5cm}
\centering
\setlength\lineskip{1.5pt}
\setlength\tabcolsep{1.5pt} 
{\small
\begin{tabular}{
c
>{\centering\arraybackslash}m{\plotheight}
>{\centering\arraybackslash}m{\plotheight}
>{\centering\arraybackslash}m{\plotheight}
}
&Seen object & Unseen object & After masking\\
\rotatebox{90}{\cite{balntas-iccv17-poseguidedrgbdfeaturelearning}}&
\includegraphics[height=\plotheight, ]{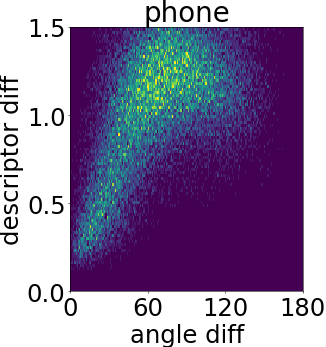}&
\includegraphics[height=\plotheight, ]{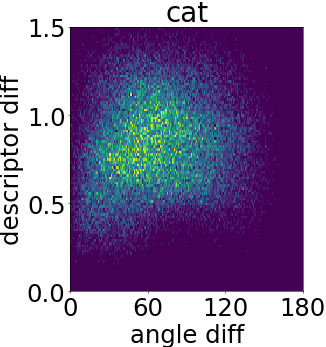}&
\includegraphics[height=\plotheight, ]{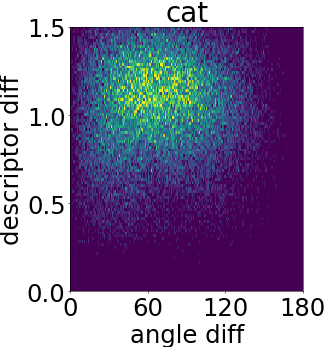}\\
\rotatebox{90}{$\!\!\!\!\!\!\!\!\!\!\!\!$Our method }&
\includegraphics[height=\plotheight, ]{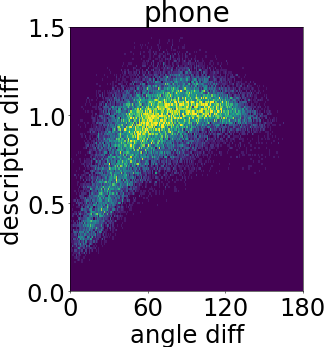}&
\includegraphics[height=\plotheight, ]{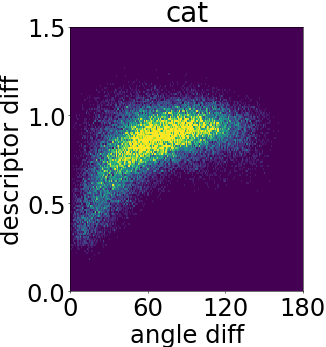}&
\includegraphics[height=\plotheight, ]{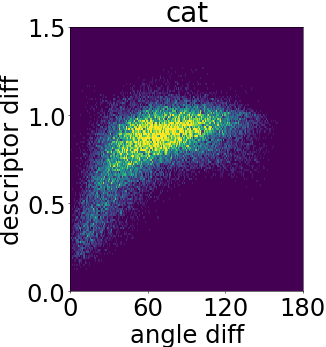}\\
\end{tabular}
}
\caption{
  \textbf{Visualization of the correlation between pose distances and representation distances, } for understanding the discriminative power of different image representations for pose retrieval. First row is for \cite{balntas-iccv17-poseguidedrgbdfeaturelearning}, second row is for our method. Please see Section~\ref{sec:pose_discrimination}.}
\label{fig:correlation}
\end{figure}

As discussed in Section~3.1.2 of the main paper, a drawback of global representations is their poor reliability to represent the real image of a new object even when the object identity is known and the background is uniform. To illustrate this, we show in Figure~\ref{fig:correlation} the correlation between pose distances and representation distances as in \cite{wohlhart-cvpr15-learningdescriptors,balntas-iccv17-poseguidedrgbdfeaturelearning}.
\cite{wohlhart-cvpr15-learningdescriptors,balntas-iccv17-poseguidedrgbdfeaturelearning} provided such plots only for seen objects and RGB-D data, we consider here objects that have been seen or unseen and we use RGB data only. As in \cite{wohlhart-cvpr15-learningdescriptors,balntas-iccv17-poseguidedrgbdfeaturelearning}, the plots of Figure~\ref{fig:correlation} are obtained by considering all possible pairs made of real images and synthetic images for a given object.

Ideally, the plots should exhibit a diagonal pattern, in the region closed to the (0, 0) point on the bottom-left of the graph. This region corresponds to the critical region for correct image/template matching. A diagonal pattern corresponds to a strong correlation between pose differences and distances between representations.

More plots are given in Section~\ref{sec:additional_result} and they all yield to the same conclusion: 
\begin{itemize}
    \item The first column of Figure~\ref{fig:correlation} shows that both representations result in a strong correlation for an seen object. 
    \item The second column shows this correlation is lost when considering a new object for the global representation but not with ours. 
\item  To check if this was due to the presence of clutter in the background of the real images, we removed the background by using the ground truth mask of the objects. The third column of Figure~\ref{fig:correlation} shows that even without background, the correlation is still very poor for global representations. This can be explained by the fact that the pooling layers remove important information for unseen objects. We postulate that the rest of the architecture, in particular the fully connected layers learns to compensate this loss of information for seen objects, but such compensation is not possible for unseen objects.
\end{itemize}

\begin{table*}[!t]
\addtolength{\tabcolsep}{-1pt}
\centering
    \scalebox{.73}{
    \begin{tabular}{l l l l| 
    c  c  c | c |
    c  c  c | c |
    c  c  c | c |
    c  c  c | c }
	\toprule
	\multirow{2}{*}{\bf Method} &
	\multirow{2}{*}{\bf Backbone} &
	\multirow{2}{*}{\bf Features} &
	\multirow{2}{*}{\bf Loss} &
	\multicolumn{4}{c|}{\textbf{Seen LM}}&
	\multicolumn{4}{c|}{\textbf{Seen O-LM}}&
	\multicolumn{4}{c|}{\textbf{Unseen LM}}& 
	\multicolumn{4}{c}{\textbf{Unseen O-LM}}\\
	
    \cmidrule(lr){5-8} \cmidrule(lr){9-12}\cmidrule(lr){13-16}\cmidrule(lr){17-20}
	
	& & & & ~\#1 & ~\#2 & ~\#3 & Avg. & ~\#1 & ~\#2 & ~\#3 & Avg. & ~\#1 & ~\#2 & ~\#3 & Avg. & ~\#1 & ~\#2 & ~\#3 & Avg. \\
	\midrule
	
	\rowcolor{gray!10}
	\cite{wohlhart-cvpr15-learningdescriptors} & Base~\cite{wohlhart-cvpr15-learningdescriptors} & Global & \cite{wohlhart-cvpr15-learningdescriptors} & 12.1 & 13.2 & 12.0 & 12.4 & 49.5 & 51.1 & 52.3&  50.9 & 54.6 & 55.7 & 59.0  & 56.4 & 59.4&  57.2&  56.0 & 57.5 \\
	
	\cite{wohlhart-cvpr15-learningdescriptors} & Base~\cite{wohlhart-cvpr15-learningdescriptors} & Global & InfoNCE~\cite{Oord2018infoNCE} & 6.6  & 6.5 & 6.7 & 6.6 & 47.9 & 45.2&  52.9& 58.6 & 58.6&  48.5&  48.1& 51.7 & 61.4 & 56.3 & 54.6& 57.3\\
	
	\rowcolor{gray!10}
	\cite{balntas-iccv17-poseguidedrgbdfeaturelearning} & Base~\cite{wohlhart-cvpr15-learningdescriptors}& Global & \cite{balntas-iccv17-poseguidedrgbdfeaturelearning} & 11.2 & 11.8 & 12.9 & 12.0 & 49.5 & 51.1 & 52.3 & 51.0 & 54.6 & 55.7 & 59.0 & 56.4 & 60.0  & 53.8 & 60.1& 57.9\\
	
	\cite{balntas-iccv17-poseguidedrgbdfeaturelearning} & Base~\cite{wohlhart-cvpr15-learningdescriptors}& Global & InfoNCE~\cite{Oord2018infoNCE} & 6.4 & 6.4 &  6.5 & 6.4 & 46.6 & 47.2 & 50.4 & 48.0 & 67.9 & 48.6 & 50.8 & 55.7 & 73.4 & 56.1 & 53.4 & 60.9\\
	
	\rowcolor{gray!10}
	Ours & Base~\cite{wohlhart-cvpr15-learningdescriptors}& Local & \cite{wohlhart-cvpr15-learningdescriptors} & 15.2 & 15.8 & 14.9 & 15.3 & 32.6 & 31.9 & 31.0 & 31.8 & 27.1 & 27.4 & 25.3 & 26.5 & 41.5 & 41.2 & 42.3 & 41.6\\
    
    Ours & Base~\cite{wohlhart-cvpr15-learningdescriptors}& Local & InfoNCE~\cite{Oord2018infoNCE} & 4.8 & 5.1 & 7.9 & 5.9 & 12.3 & 18.5& 21.8 & 18.5 & 15.4 & 9.9 & 20.3 & 15.2 & 32.3 & 21.3 & 17.6 & 23.7\\
    
    
    \cmidrule(lr){1-20}
    
    \rowcolor{gray!10}
	
	\cite{wohlhart-cvpr15-learningdescriptors} & ResNet50 ~\cite{he-cvpr16-deepresiduallearning} & Global & InfoNCE ~\cite{Oord2018infoNCE} & 3.6 & \bf 4.3 & 4.7 & 4.2  & 26.7 & 29.8 & 34.5  & 30.3 & 43.1 & 42.7 & 40.5 & 42.1 & 45.8 & 51.8 & 44.0 & 47.1\\
	
	\cite{balntas-iccv17-poseguidedrgbdfeaturelearning} & ResNet50 ~\cite{he-cvpr16-deepresiduallearning}& Global & InfoNCE ~\cite{Oord2018infoNCE}  & 3.7 & 4.5 & 5.1 & 4.4 & 35.7 & 29.8 & 39.7 & 35.1 & 51.1 &  50.0 & 39.3 & 46.8  & 64.5  & 61.0 & 49.8 & 58.4\\
    
    \rowcolor{gray!10}
    Ours & ResNet50 ~\cite{he-cvpr16-deepresiduallearning}& Local & InfoNCE ~\cite{Oord2018infoNCE} & \bf 3.3 & 4.6 & \bf 3.4 & \bf 3.7 &  \bf 9.7 & \bf 11.1 &  \bf 11.5 &  \bf 10.7 & \bf 7.5 & \bf 3.1 & \bf 10.0 & \bf 6.9 & \bf 17.5 & \bf 11.5 & \bf 7.5 & \bf 12.2
\\
\bottomrule
\end{tabular}}

\vspace*{-2.5mm}
\caption{\textbf{Comparison of our method with \cite{ wohlhart-cvpr15-learningdescriptors} and \cite{balntas-iccv17-poseguidedrgbdfeaturelearning}} on seen and unseen objects of LINEMOD~(LM) and Occlusion-LINEMOD~(O-LM) for the three splits detailed in Section 4.1 of the main paper. We report here the pose error, measured by the angle between the positions on the half-sphere for the ground truth pose and the predicted pose.}

\label{tab:linemod_full_err}
\end{table*}

\section{Training details}
\label{sec:training_details}
\begin{figure}[!t]
    \begin{center}
    \includegraphics[width=1\linewidth]{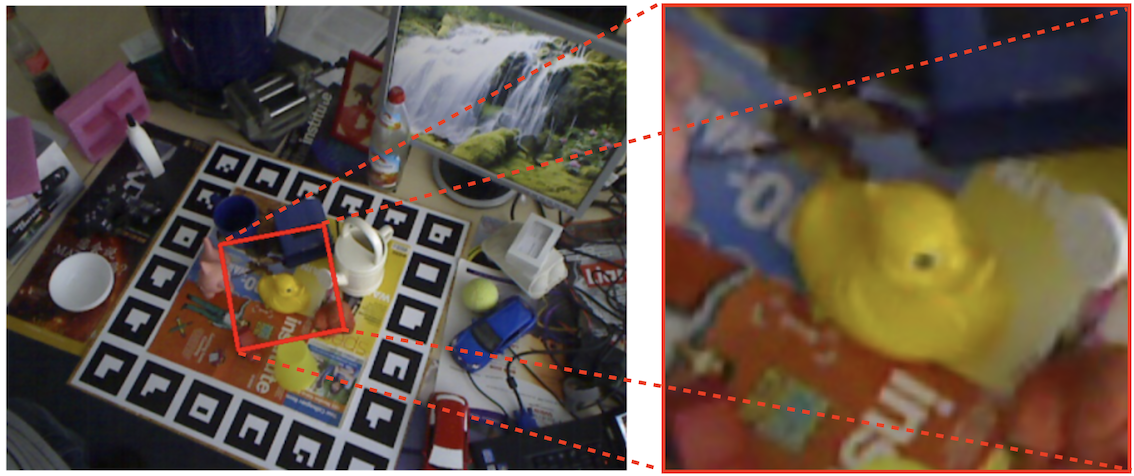}
    \end{center}
    \vspace{-6mm}
    \caption{
        \label{fig:cropping}
        {\bf Cropping on LINEMOD.} the cropping on LINEMOD and Occlusion-LINEMOD is done by virtually setting a box centered at the object. Please see Section \ref{sec:training_details}.}
\end{figure}

\paragraph{Cropping on LINEMOD.} Unless otherwise stated in previous works~\cite{wohlhart-cvpr15-learningdescriptors,balntas-iccv17-poseguidedrgbdfeaturelearning}, the cropping on LINEMOD and Occlusion-LINEMOD is done by virtually setting a box, 40 cm in each dimension, centered at the object as shown in Figure \ref{fig:cropping}. When
all the patches are extracted, we normalize them to the desired crop size. Please note that with this cropping, we do not consider in-plane rotations, in other words, we omit one additional degree of freedom.

\begin{figure}[!t]
\setlength\plotheight{2.0cm}
\centering
\setlength\lineskip{1.5pt}
\setlength\tabcolsep{1.5pt} 
{\small
\begin{tabular}{
>{\centering\arraybackslash}m{\plotheight}
>{\centering\arraybackslash}m{\plotheight}
>{\centering\arraybackslash}m{\plotheight}
>{\centering\arraybackslash}m{\plotheight}
}
\includegraphics[height=\plotheight, ]{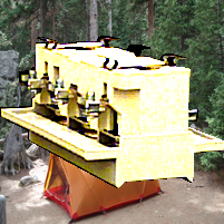}&
\includegraphics[height=\plotheight, ]{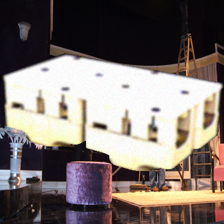}&
\includegraphics[height=\plotheight, ]{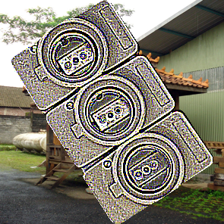}&
\includegraphics[height=\plotheight, ]{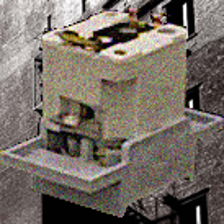}\\
\end{tabular}
}
\vspace{-2mm}
\caption{\textbf{Examples of training images.} We use randomized background from SUN397 \cite{xiao2010sun} with data augmentation, including Gaussian blur, contrast, brightness, color and sharpness filters from the Pillow library \cite{pillow}.}
\label{fig:training_images}
\end{figure}

\paragraph{Data augmentation on T-LESS.} As done in \cite{sundermeyer-eccv18-implicit3dorientationlearning, sundermeyer-cvpr20-multipathlearning}, we also apply data augmentation to the input images of T-LESS during training. We use Gaussian blur, contrast, brightness, color, and sharpness filters with the Pillow library~\cite{pillow}. Some training samples can be seen in Figure~\ref{fig:training_images}.

\paragraph{Pre-trained Features.} We intialize the network ResNet50 with MOCOv2's features \cite{chen2020mocov2}. It has been shown in \cite{Xiao2020PoseContrast} that this initialization can improve both the convergence and the performance. We show in Table \ref{tab:moco_comparison} our comparison on T-LESS dataset when training the network ResNet50 from scratch and with initializing pre-trained features of MOCOv2 \cite{chen2020mocov2}.
\begin{table}[t]
	\centering
    \resizebox{0.90\linewidth}{!}{
	\begin{tabular}{@{}l c c c c@{}}
	\toprule
    \multirow{2}{*}{\bf Initialization} & \multirow{2}{*}{\bf \parbox{1.5cm}{Number templates}} & 
    \multicolumn{3}{c}{\textbf{Recall VSD}}\\
	\cmidrule(lr){3-5}
	& &\textbf{Obj. 1-18} & \textbf{Obj. 19-30} & \textbf{Avg} \\
	\midrule
	\rowcolor{gray!10} From scratch & 21K & 55.42 & 51.40 & 53.81\\
	MOCOv2 \cite{chen2020mocov2} & 21K & \bf 59.14 & \bf 56.91 & \bf 58.25\\
	
	\bottomrule
	\end{tabular}}
	\vspace{-2mm}	
	\caption{{\bf Network initializations evaluated on T-LESS}. Using pre-trained features from MOCOv2 \cite{chen2020mocov2} brings some improvement comparing to training from scratch.}
    \label{tab:moco_comparison}
\end{table}
\section{Projective distance estimation}
\label{sec:translation_tless}
As done in \cite{sundermeyer-eccv18-implicit3dorientationlearning, sundermeyer-cvpr20-multipathlearning}, we estimate 3D translation in the query image from the retrieved template and the input bounding box as detailed in Section 3.6.2 of \cite{sundermeyer-ijcv20-augmentedautoencoders}. More precisely, given known camera intrinsic of both real sensor $K_{query}$ and of the synthetic view $K_{temp}$, we estimate the distance $\hat{t}_{query, z}$ of real image:
\begin{equation}
    \hat{t}_{query, z} = t_{temp, z} \times \frac{||bb_{temp}||}{||bb_{query}||} \times \frac{f_{query}}{f_{temp}} 
\label{loss:translationz}
\end{equation}
where $||bb_{(.)}||$ is the diagonal of the bounding box and $||f_{(.)}||$ is the focal length.

Then, we can estimate the vector to transform from the object center in the synthetic view to the query image:
\begin{equation}
    \Delta \hat{t} = \hat{t}_{query, z} K^{-1}_{query} bb_{query, c} - \hat{t}_{temp, z} K^{-1}_{temp} bb_{temp, c}
\end{equation}
where $bb_{(.), c}$ is the bounding box centers in homogeneous coordinates.

Finally, the 3D translation in the query image $\hat{t}_{query}$ can be estimated as :
\begin{equation}
    \hat{t}_{query} = \hat{t}_{temp} + \Delta \hat{t}
\end{equation}
where $\hat{t}_{temp}=(0, 0, \hat{t}_{temp, z})$, the translation from camera to object center in the synthetic view.
\section{Additional Results}
\label{sec:additional_result}

\subsection{Quantitative Results}
\label{quantitative}
We show in Table~\ref{tab:linemod_full_err} quantitative results with pose error, measured by the angle between the two positions on the viewing half-sphere. 

\subsection{Qualitative Results}
\label{qualitative}
\begin{figure*}
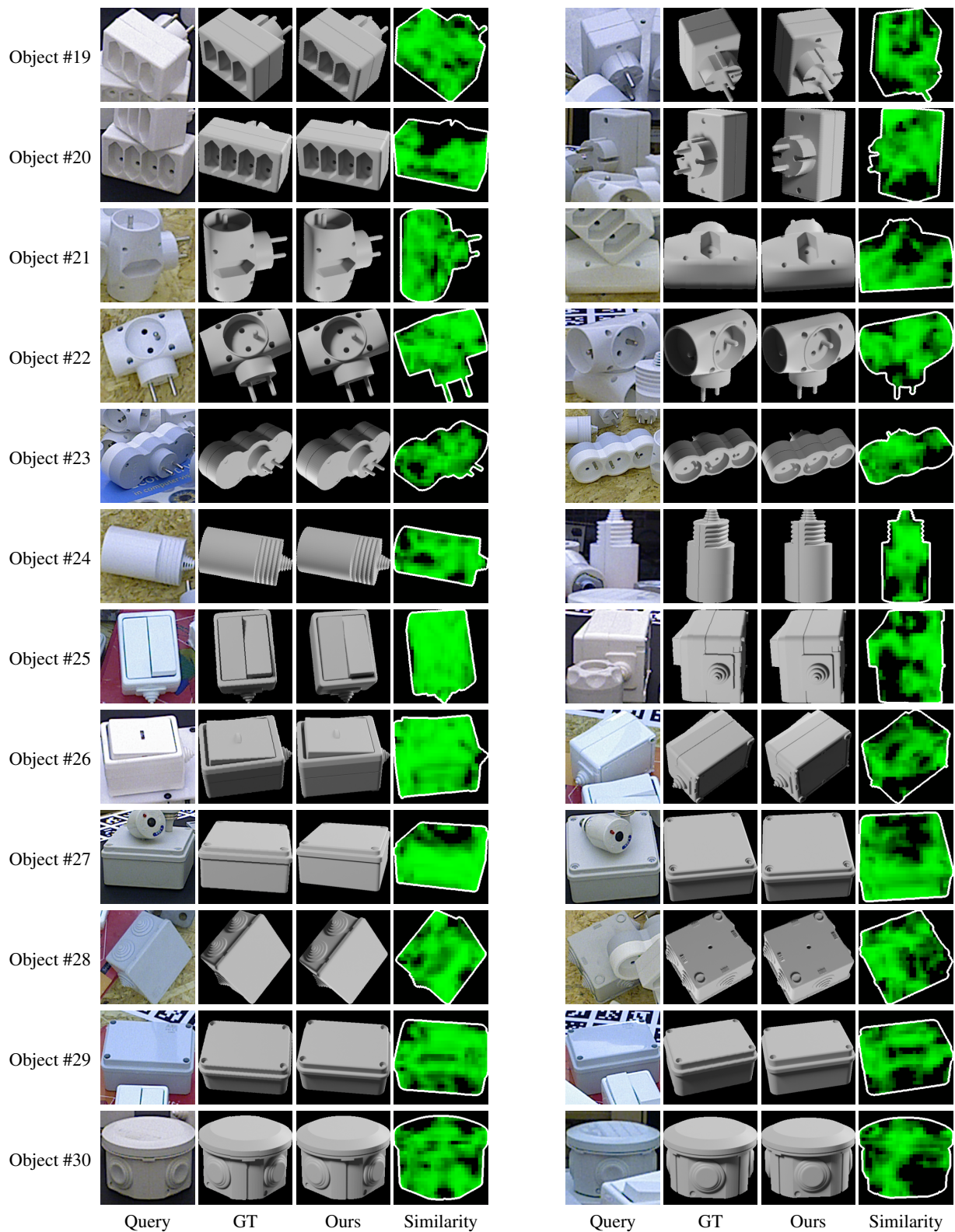

\setlength\imageheight{1.6cm}
\setlength\plotheight{2.5cm}
\centering
\setlength\lineskip{1.pt}
\setlength\tabcolsep{1.pt} 
{\small
\\
}
\caption{{\bf Qualitative results on unseen objects of T-LESS dataset.}}
\label{fig:tless}
\end{figure*}

\begin{figure*}[!t]
\setlength\plotheight{1.85cm}
\centering
\setlength\lineskip{0.5pt}
\setlength\tabcolsep{0.5pt} 
{\small
\begin{tabular}{
c
>{\centering\arraybackslash}m{0.15cm}
>{\centering\arraybackslash}m{1.0cm}
>{\centering\arraybackslash}m{\plotheight}
>{\centering\arraybackslash}m{\plotheight}
>{\centering\arraybackslash}m{\plotheight}
>{\centering\arraybackslash}m{\plotheight}
>{\centering\arraybackslash}m{\plotheight}
>{\centering\arraybackslash}m{\plotheight}
>{\centering\arraybackslash}m{\plotheight}
>{\centering\arraybackslash}m{\plotheight}
>{\centering\arraybackslash}m{\plotheight}
>{\centering\arraybackslash}m{\plotheight}
}
\rotatebox{90}{\multirow{2}{*}{\bf$\!\!\!\!\!\!\!\!\!\!\!\!\!\!\!\!\!\!\!\!\!\!\!\!\!\!$Seen}}& &
\rotatebox{90}{\cite{balntas-iccv17-poseguidedrgbdfeaturelearning}}&
\includegraphics[height=\plotheight, ]{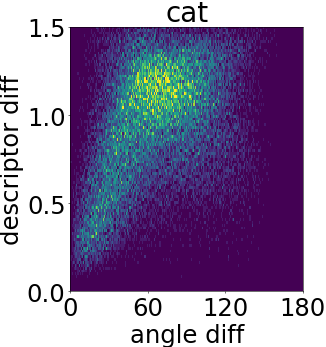}&
\includegraphics[height=\plotheight, ]{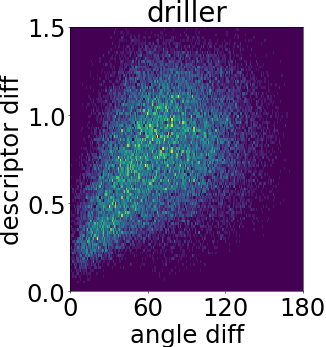}&
\includegraphics[height=\plotheight, ]{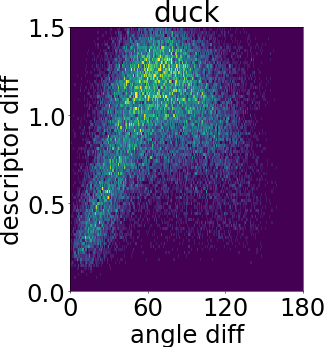}&
\includegraphics[height=\plotheight, ]{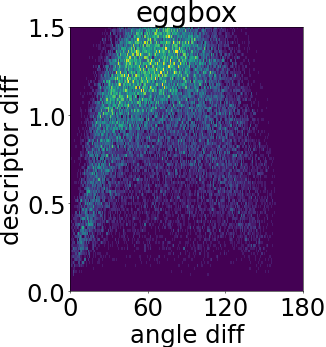}&
\includegraphics[height=\plotheight, ]{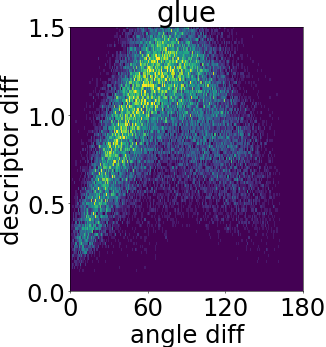}&
\includegraphics[height=\plotheight, ]{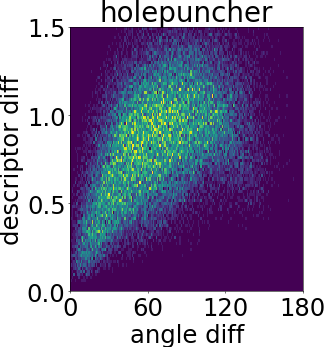}&
\includegraphics[height=\plotheight, ]{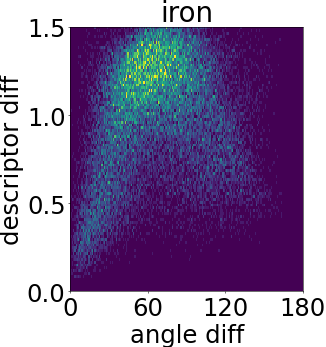}\\
\end{tabular}

\begin{tabular}{
c
>{\centering\arraybackslash}m{0.15cm}
>{\centering\arraybackslash}m{1.0cm}
>{\centering\arraybackslash}m{\plotheight}
>{\centering\arraybackslash}m{\plotheight}
>{\centering\arraybackslash}m{0.7cm}
c
>{\centering\arraybackslash}m{0.15cm}
>{\centering\arraybackslash}m{1.0cm}
>{\centering\arraybackslash}m{\plotheight}
>{\centering\arraybackslash}m{\plotheight}
>{\centering\arraybackslash}m{\plotheight}
>{\centering\arraybackslash}m{\plotheight}
>{\centering\arraybackslash}m{\plotheight}
}
\rotatebox{90}{\multirow{2}{*}{\bf$\!\!\!\!\!\!\!\!\!\!\!\!\!\!\!\!\!\!\!\!\!\!\!\!\!\!$Seen}}& &
\rotatebox{90}{\cite{balntas-iccv17-poseguidedrgbdfeaturelearning}}& 
\includegraphics[height=\plotheight, ]{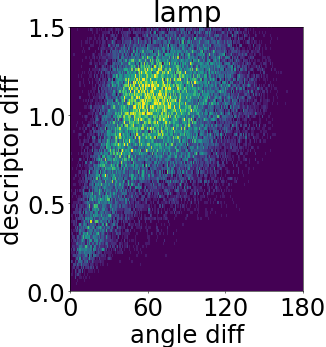}&
\includegraphics[height=\plotheight, ]{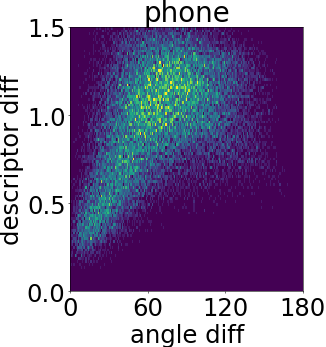}&
& \rotatebox{90}{\multirow{2}{*}{\bf$\!\!\!\!\!\!\!\!\!\!\!\!\!\!\!\!\!\!\!\!\!\!\!\!\!\!$Unseen}}&
&\rotatebox{90}{\cite{balntas-iccv17-poseguidedrgbdfeaturelearning}}& 
\includegraphics[height=\plotheight, ]{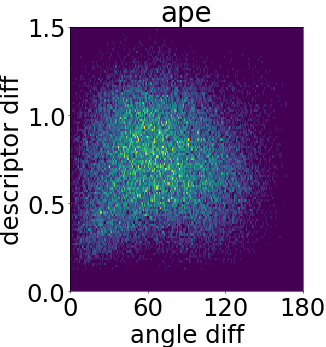}&
\includegraphics[height=\plotheight, ]{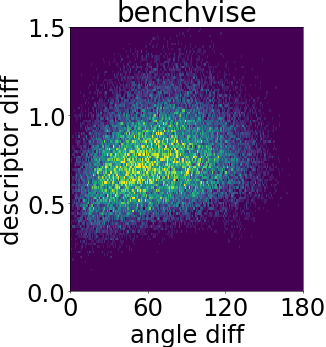}&
\includegraphics[height=\plotheight, ]{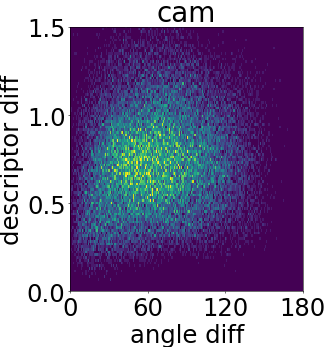}&
\includegraphics[height=\plotheight, ]{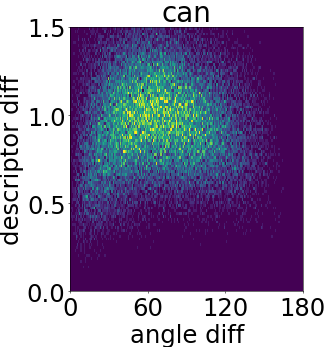}\\
\end{tabular}
}
\vspace{-3mm}
\caption{\textbf{Visualization of the correlation between pose distances and representation distances on Split~\#1}: of unseen objects of LINEMOD (two first rows and two first columns from the left of two last rows) and unseen objects of LINEMOD (fours last columns from the left of two last rows). Ideally, the plots should exhibit the diagonal pattern at the region closed to the (0, 0) point on the bottom-left that corresponds to the critical region for correct image/template matching, showing a strong correlation between pose differences and representation distances. The plots of seen objects of LINEMOD show that both representations result in a strong correlation for training objects. The plots of unseen objects of LINEMOD show this correlation is lost when considering a new object for the global representation~\cite{balntas-iccv17-poseguidedrgbdfeaturelearning} but not with ours.}
\label{fig:descp_split1}
\end{figure*}

\begin{figure*}[!t]
\setlength\plotheight{1.85cm}
\centering
\setlength\lineskip{0.5pt}
\setlength\tabcolsep{0.5pt} 
{\small
\begin{tabular}{
c
>{\centering\arraybackslash}m{0.15cm}
>{\centering\arraybackslash}m{1.0cm}
>{\centering\arraybackslash}m{\plotheight}
>{\centering\arraybackslash}m{\plotheight}
>{\centering\arraybackslash}m{\plotheight}
>{\centering\arraybackslash}m{\plotheight}
>{\centering\arraybackslash}m{\plotheight}
>{\centering\arraybackslash}m{\plotheight}
>{\centering\arraybackslash}m{\plotheight}
>{\centering\arraybackslash}m{\plotheight}
>{\centering\arraybackslash}m{\plotheight}
>{\centering\arraybackslash}m{\plotheight}
}
\rotatebox{90}{\multirow{2}{*}{\bf$\!\!\!\!\!\!\!\!\!\!\!\!\!\!\!\!\!\!\!\!\!\!\!\!\!\!$Seen}}& &
\rotatebox{90}{\cite{balntas-iccv17-poseguidedrgbdfeaturelearning}}&
\includegraphics[height=\plotheight, ]{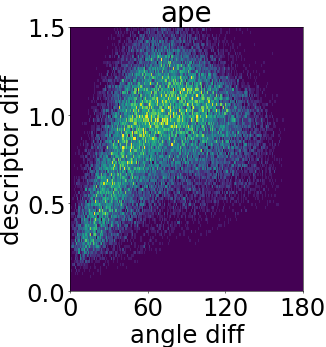}&
\includegraphics[height=\plotheight, ]{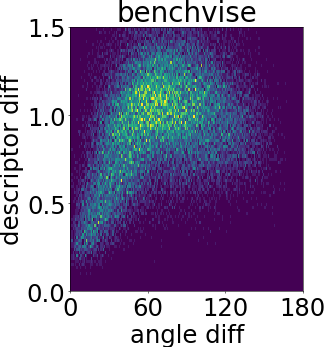}&
\includegraphics[height=\plotheight, ]{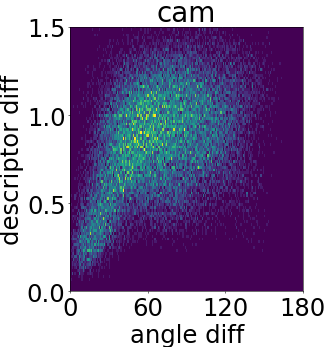}&
\includegraphics[height=\plotheight, ]{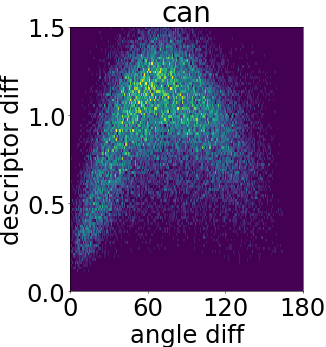}&
\includegraphics[height=\plotheight, ]{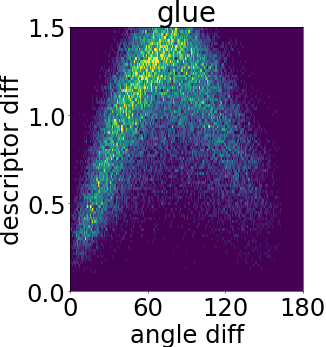}&
\includegraphics[height=\plotheight, ]{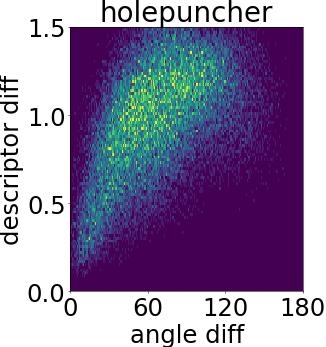}&
\includegraphics[height=\plotheight, ]{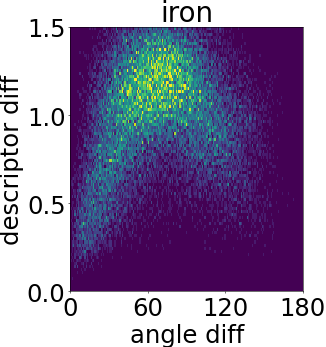}\\
\end{tabular}

\begin{tabular}{
c
>{\centering\arraybackslash}m{0.15cm}
>{\centering\arraybackslash}m{1.0cm}
>{\centering\arraybackslash}m{\plotheight}
>{\centering\arraybackslash}m{\plotheight}
>{\centering\arraybackslash}m{0.7cm}
c
>{\centering\arraybackslash}m{0.15cm}
>{\centering\arraybackslash}m{1.0cm}
>{\centering\arraybackslash}m{\plotheight}
>{\centering\arraybackslash}m{\plotheight}
>{\centering\arraybackslash}m{\plotheight}
>{\centering\arraybackslash}m{\plotheight}
>{\centering\arraybackslash}m{\plotheight}
}
\rotatebox{90}{\multirow{2}{*}{\bf$\!\!\!\!\!\!\!\!\!\!\!\!\!\!\!\!\!\!\!\!\!\!\!\!\!\!$Seen}}& &
\rotatebox{90}{\cite{balntas-iccv17-poseguidedrgbdfeaturelearning}}& 
\includegraphics[height=\plotheight, ]{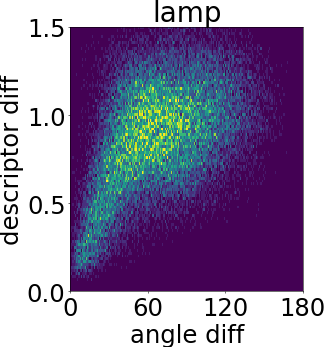}&
\includegraphics[height=\plotheight, ]{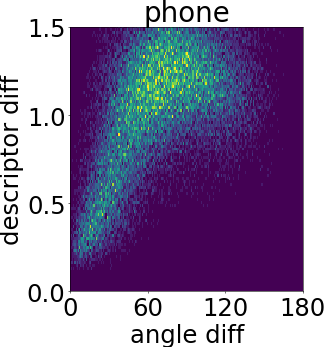}&
& \rotatebox{90}{\multirow{2}{*}{\bf$\!\!\!\!\!\!\!\!\!\!\!\!\!\!\!\!\!\!\!\!\!\!\!\!\!\!$Unseen}}&
&\rotatebox{90}{\cite{balntas-iccv17-poseguidedrgbdfeaturelearning}}& 
\includegraphics[height=\plotheight, ]{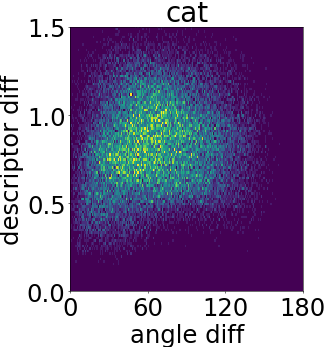}&
\includegraphics[height=\plotheight, ]{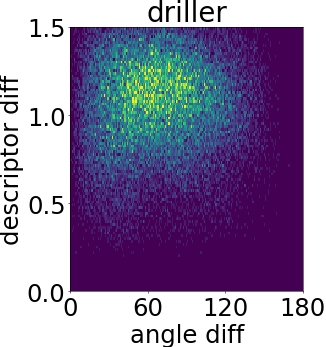}&
\includegraphics[height=\plotheight, ]{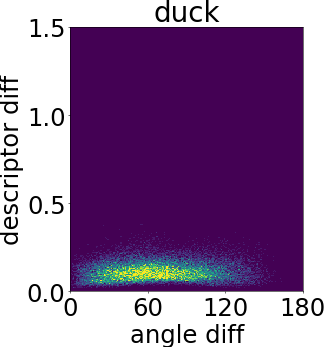}&
\includegraphics[height=\plotheight, ]{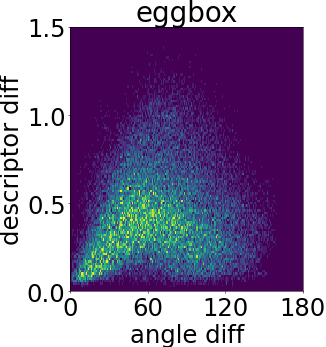}\\
\end{tabular}
}
\caption{\textbf{Visualization of the correlation between pose distances and representation distances on Split~\#2}: seen objects of LINEMOD (two first rows and two first columns from the left of two last rows) and unseen objects of LINEMOD (fours last columns from the left of two last rows).}

\label{fig:descp_split2}
\end{figure*}

\begin{figure*}[!t]
\setlength\plotheight{1.95cm}
\centering
\setlength\lineskip{0.5pt}
\setlength\tabcolsep{0.5pt} 
{\small
\begin{tabular}{
c
>{\centering\arraybackslash}m{0.15cm}
>{\centering\arraybackslash}m{1.0cm}
>{\centering\arraybackslash}m{\plotheight}
>{\centering\arraybackslash}m{\plotheight}
>{\centering\arraybackslash}m{\plotheight}
>{\centering\arraybackslash}m{\plotheight}
>{\centering\arraybackslash}m{\plotheight}
>{\centering\arraybackslash}m{\plotheight}
>{\centering\arraybackslash}m{\plotheight}
>{\centering\arraybackslash}m{\plotheight}
>{\centering\arraybackslash}m{\plotheight}
>{\centering\arraybackslash}m{\plotheight}
>{\centering\arraybackslash}m{\plotheight}
}
\rotatebox{90}{\multirow{2}{*}{\bf$\!\!\!\!\!\!\!\!\!\!\!\!\!\!\!\!\!\!\!\!\!\!\!\!\!\!$Seen}}& &
\rotatebox{90}{\cite{balntas-iccv17-poseguidedrgbdfeaturelearning}}&
\includegraphics[height=\plotheight, ]{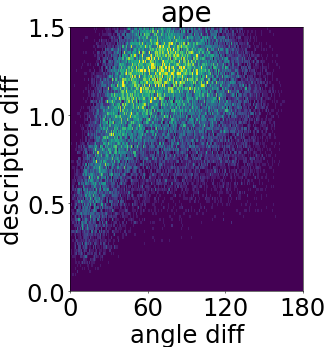}&
\includegraphics[height=\plotheight, ]{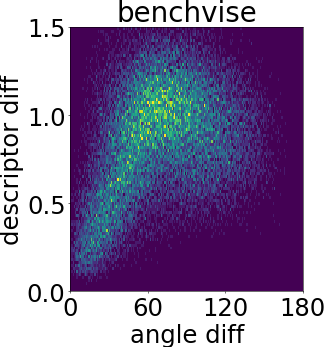}&
\includegraphics[height=\plotheight, ]{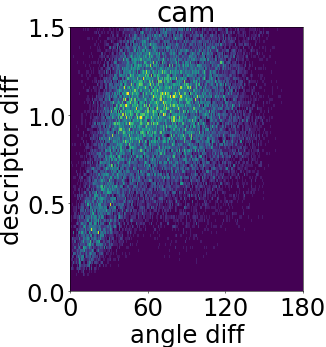}&
\includegraphics[height=\plotheight, ]{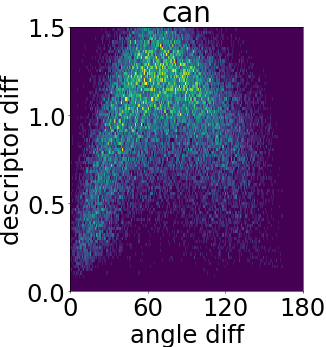}&
\includegraphics[height=\plotheight, ]{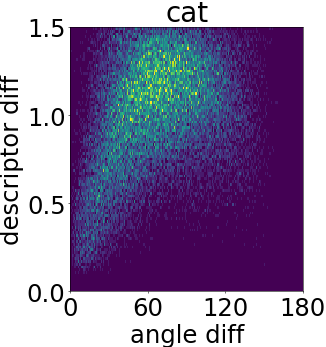}&
\includegraphics[height=\plotheight, ]{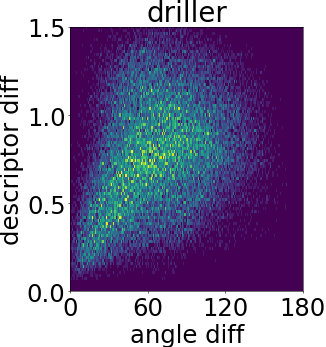}&
\includegraphics[height=\plotheight, ]{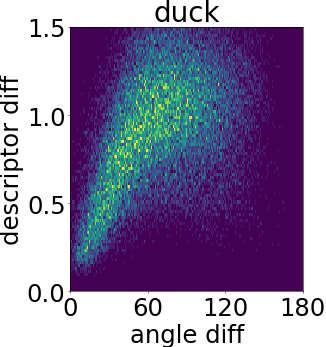}&
\includegraphics[height=\plotheight, ]{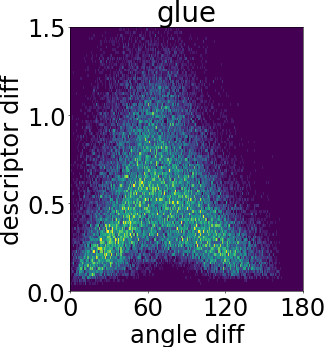}\\
\end{tabular}

\begin{tabular}{
c
>{\centering\arraybackslash}m{0.15cm}
>{\centering\arraybackslash}m{1.0cm}
>{\centering\arraybackslash}m{\plotheight}
>{\centering\arraybackslash}m{\plotheight}
>{\centering\arraybackslash}m{\plotheight}
>{\centering\arraybackslash}m{\plotheight}
>{\centering\arraybackslash}m{\plotheight}
>{\centering\arraybackslash}m{\plotheight}
>{\centering\arraybackslash}m{\plotheight}
>{\centering\arraybackslash}m{\plotheight}
>{\centering\arraybackslash}m{\plotheight}
>{\centering\arraybackslash}m{\plotheight}
>{\centering\arraybackslash}m{\plotheight}
>{\centering\arraybackslash}m{\plotheight}
}
\rotatebox{90}{\multirow{2}{*}{\bf$\!\!\!\!\!\!\!\!\!\!\!\!\!\!\!\!\!\!\!\!\!\!\!\!\!\!$Unseen}}& &
\rotatebox{90}{\cite{balntas-iccv17-poseguidedrgbdfeaturelearning}}&
\includegraphics[height=\plotheight, ]{images/supp_mat/LM_split3_descriptors_balnas/split3_descp_unseen_glue_balnas.png}&
\includegraphics[height=\plotheight, ]{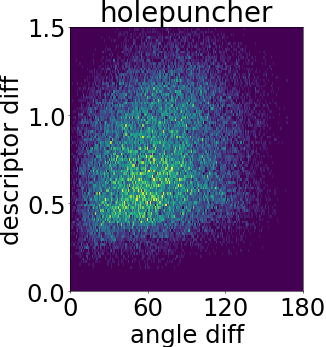}&
\includegraphics[height=\plotheight, ]{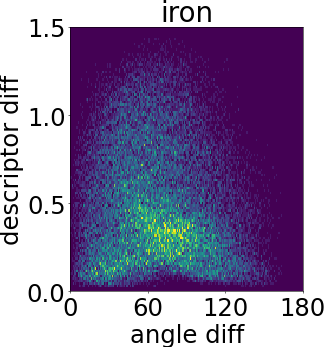}&
\includegraphics[height=\plotheight, ]{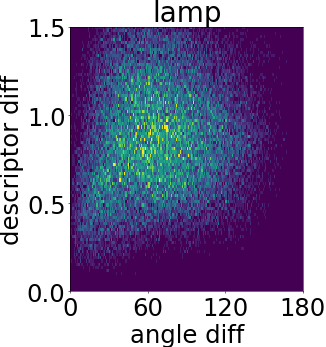}&
\includegraphics[height=\plotheight, ]{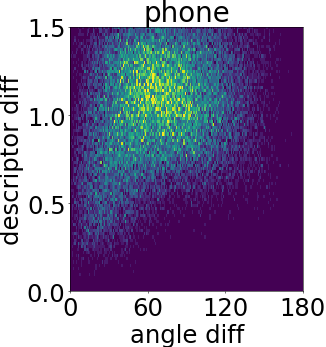}\\
\end{tabular}
}
\caption{\textbf{Visualization of the correlation between pose distances and representation distances on Split~\#3}: seen objects of LINEMOD (two first rows and two first columns from the left of two last rows) and unseen objects of LINEMOD (fours last columns from the left of two last rows). }
\label{fig:descp_split3}
\end{figure*}

We show in Figures \ref{fig:tless}, \ref{fig:split1}, \ref{fig:descp_split1}, \ref{fig:split2}, \ref{fig:descp_split2}, \ref{fig:split3}, \ref{fig:descp_split3} additional qualitative results on T-LESS and each split of LINEMOD and Occlusion-LINEMOD.
\newpage
\newpage
\begin{figure*}
\setlength\imageheight{1.35cm}

\setlength\plotheight{2.5cm}
\centering
\setlength\lineskip{1.pt}
\setlength\tabcolsep{1.pt} 
{\small
%
}
\caption{{\bf Qualitative results four test sets on Split~\#1:} of seen objects of LINEMOD (top left), seen objects of Occlusion-LINEMOD (top right), unseen objects of LINEMOD (bottom left) and unseen objects of Occlusion-LINEMOD (bottom right).}
\label{fig:split1}
\end{figure*}
\newpage
\begin{figure*}
\setlength\imageheight{1.35cm}

\setlength\plotheight{2.5cm}
\centering
\setlength\lineskip{1.pt}
\setlength\tabcolsep{1.pt} 
{\small

}
\caption{{\bf Qualitative results four test sets on Split~\#2:} of seen objects of LINEMOD (top left), seen objects of Occlusion-LINEMOD (top right), unseen objects of LINEMOD (bottom left) and unseen objects of Occlusion-LINEMOD (bottom right).}
\label{fig:split2}
\end{figure*}
\begin{figure*}
\setlength\imageheight{1.35cm}

\setlength\plotheight{2.5cm}
\centering
\setlength\lineskip{1.pt}
\setlength\tabcolsep{1.pt} 
{\small
%
}
\caption{{\bf Qualitative results four test sets on Split~\#3:} of seen objects of LINEMOD (top left), seen objects of Occlusion-LINEMOD (top right), unseen objects of LINEMOD (bottom left) and unseen objects of Occlusion-LINEMOD (bottom right).}
\label{fig:split3}
\end{figure*}
\end{document}